\documentclass[10pt,twocolumn,letterpaper]{article}

\PassOptionsToPackage{table}{xcolor}

\usepackage[pagenumbers]{cvpr} % arXiv version: authors shown, page numbers on

\usepackage[T1]{fontenc}

\usepackage{multirow}

\providecommand{\vn}[1]{{\scriptsize(#1)}}

\definecolor{tabhl}{RGB}{214,232,246}   % "ours" rows
\definecolor{tabhl2}{RGB}{228,240,249}  % runner-up row (the baseline just above)
\definecolor{tabred}{RGB}{190,30,45}    % delta-vs-best numbers

\newsavebox{\fitbox}
\newcommand{\fitwidth}[1]{%
  \sbox{\fitbox}{#1}%
  \ifdim\wd\fitbox>\linewidth
    \resizebox{\linewidth}{!}{\usebox{\fitbox}}%
  \else
    \usebox{\fitbox}%
  \fi}

\definecolor{cvprblue}{rgb}{0.21,0.49,0.74}
\usepackage[pagebackref,breaklinks,colorlinks,allcolors=cvprblue]{hyperref}

\def\confName{CVPR}
\def\confYear{2027}

\title{One Affinity, Two Axes: Bidirectional Slot Attention for Iterative
Aggregation across the Structure--Detail Spectrum in Visual Retrieval}

\author{Zihang Wang$^{1}$, Wenkai Zhu$^{1}$, Lingdong Kong$^{2}$, Teng Yan$^{3}$, Alan Liang$^{4}$\\
Shuyuan Li$^{5}$, Xieyuanli Chen$^{6}$, Dong Kong$^{5}$, Guanyu Zong$^{1,2}$\\
\small $^{1}$Southeast University \quad $^{2}$National University of Singapore\\
\small $^{3}$The Hong Kong University of Science and Technology (Guangzhou)\\
\small $^{4}$Shanghai Innovation Institute \quad $^{5}$Shandong University of Science and Technology\\
\small $^{6}$National University of Defense Technology}

\begin{document}
\maketitle

\begin{abstract}
Aggregating dense patch tokens into compact and discriminative global descriptors is central to visual retrieval. Although foundation models provide rich local features, existing aggregation heads still rely on single-pass mappings and lack an explicit mechanism to coordinate structural and fine-grained cues. Therefore, we propose Bidirectional Slot Attention (BiSA), which reformulates aggregation as an iterative coupled system. Cross-Axis Consensus (CAC) progressively refines learnable slots by interpreting a single shared token--slot affinity along two complementary axes. Along the competition axis, slots vie for each token under a temperature shared across all of them, determining token ownership and inducing a non-redundant scene decomposition. Along the selection axis, a slot-specific temperature governs the breadth of each slot's response, spanning a spectrum from broad layouts to sparse fine-grained details. Their consensus couples non-redundant decomposition with multi-granular slot specialization. Furthermore, the two axes expose intrinsic token-reliability cues at no extra cost: a token's claim strength under competition and its salience under selection. Progressive Token Filtering (PTF) converts these cues into a retention probability that down-weights unreliable tokens before the next CAC iteration, without any segmentation branch or external prior, forming a self-reinforcing loop of competitive decomposition, selective focusing, and feedback purification. Across all three retrieval tasks, BiSA leads on five of six place recognition benchmarks (+4.2 R@1 on Baidu Mall), tops all 16 metrics of drone--satellite geo-localization (+8.4 R@1 at 150 m), and improves instance-level retrieval on ROxford-Hard by 4.5 mAP points. Notably, the selection axis adds only 20 learnable scalars, yet improves average R@1 by 2.8 points.
\end{abstract}

\section{Introduction}

Visual retrieval aims to identify query-relevant images from large-scale databases using compact and discriminative descriptors. To do so, a descriptor must remain invariant to substantial appearance and geometric variations while discriminating against visually similar distractors. The former relies largely on stable, large-scale structural cues, whereas the latter often depends on sparse but discriminative local evidence, such as signs, entrances, decorations, and textures. This creates a fundamental tension between structural invariance and detail-level discriminability. As visual foundation models make patch tokens increasingly expressive, the principal challenge has shifted from extracting stronger local features to aggregating them into a compact descriptor that reconciles these two complementary requirements.

\begin{figure}
    \centering
    \includegraphics[width=0.98\linewidth]{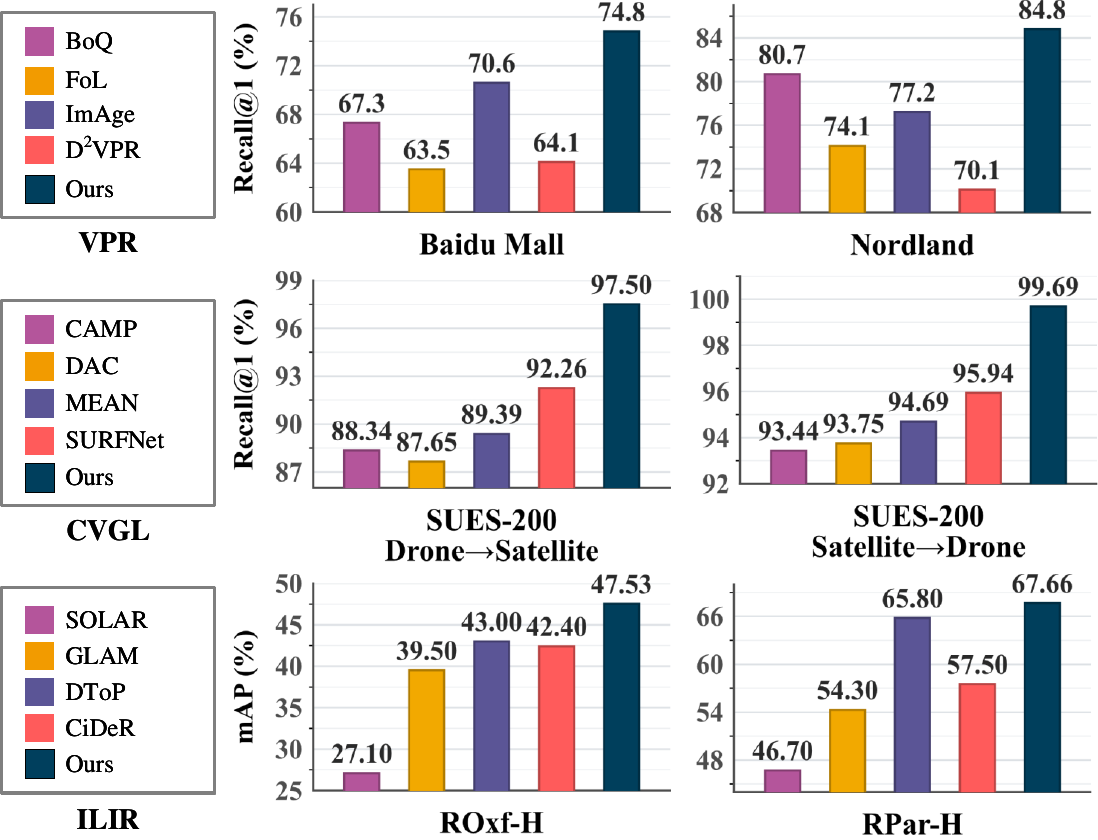}
    \caption{Cross-task comparison on representative challenging benchmarks. BiSA consistently outperforms recent methods across VPR, CVGL, and ILIR.}
    \label{fig:intro}
\end{figure}

Current aggregation heads fall short of this goal in two distinct ways. The first concerns the process of token-to-descriptor aggregation: across visual place recognition (VPR), cross-view geo-localization (CVGL), and instance-level retrieval (ILIR), dense local tokens are transformed and combined through assignment, spatial reweighting, or pooling to form the global descriptor. Representative methods instantiate this process through cluster assignment, feature mixing, or learnable queries in VPR \cite{arandjelovic2016netvlad,alibey2023mixvpr,izquierdo2024optimal,alibey2024boq}, spatially aware aggregation in CVGL \cite{shi2019spatial,liu2026surfnet}, and global pooling or object-aware selection in ILIR \cite{radenovic2018fine,ng2020solar,song2022glam,song2024cider}. Despite these different formulations, they typically fix the complex token-to-descriptor mapping within a single forward pass, resulting in a unidirectional dependency: token assignments determine descriptor components without an explicit feedback loop to progressively refine the mapping. The second concerns the granularity of the evidence retained in the descriptor. SegVLAD, StructVPR and SURFNet reinforce structural representation via SAM-generated segmentation masks, distilled structural knowledge, and surface-aware priors respectively
\cite{garg2024revisit,kirillov2023segment,shen2023structvpr,liu2026surfnet}, whereas FoL strengthens detail representation by mining reliable discriminative local regions \cite{wang2025focus}. Each of these designs, however, commits to
either structure or detail, and none explicitly models within a single descriptor the full granularity spectrum that runs from broad layouts to sparse fine-grained details.

To address both limitations, we propose Bidirectional Slot Attention (BiSA), which reformulates feature aggregation as an iterative coupled system by integrating Cross-Axis Consensus (CAC) with progressive token filtering (PTF). Within CAC, a learnable set of slots \cite{locatello2020slot} serves as an evolving aggregation state for local token features and is updated by interpreting a shared token--slot affinity along two complementary axes under asymmetric temperatures. Along the competition axis, normalization across slots under a shared temperature establishes relative token ownership, encouraging different slots to cover complementary scene content. Along the selection axis, normalization across tokens with slot-specific temperatures controls each slot's support breadth, enabling the slot set to span the structure--detail spectrum, from broad layouts to sparse discriminative cues. CAC retains token--slot associations supported by both distributions, coupling non-redundant scene decomposition with multi-granular slot specialization.

Crucially, the same axis-wise responses also expose complementary token-reliability cues at no extra cost: each token's claim strength under competition and salience under selection. PTF therefore combines these cues with the token features to suppress unreliable tokens and feeds the refined token pool back into the next CAC iteration. The next iteration thus recomputes competition and selection from a progressively cleaner token pool. Repeating this CAC--PTF feedback forms a self-reinforcing loop of competitive decomposition, selective focusing, and feedback purification. Since both decomposition and the reliability cues originate from the shared affinity, BiSA requires no external segmentation or semantic branch. As summarized in Figure~\ref{fig:intro}, BiSA consistently outperforms recent methods across VPR, CVGL, and ILIR, demonstrating its generality.

To summarize, our work brings the following contributions: \textbf{1)} We propose Bidirectional Slot Attention (BiSA), an iterative aggregation framework that coordinates structural invariance and detail-level discriminability across a continuous granularity spectrum without external priors.
\textbf{2)} We introduce Cross-Axis Consensus (CAC), which applies shared-temperature competition and slot-specific-temperature selection to a common token--slot affinity, coupling non-redundant decomposition with slot specialization across the structure--detail spectrum.
\textbf{3)} We introduce Progressive Token Filtering (PTF), which combines the axis-wise reliability responses with token features to predict per-token retention probabilities without additional supervision. The reweighted tokens re-enter CAC, coupling slot refinement with token filtering in a single evolving system.
\textbf{4)} The same aggregation architecture consistently leads across VPR, CVGL, and ILIR, including zero-shot cross-view transfer without target fine-tuning, supporting iterative structure--detail consensus as a task-general retrieval principle.

% \begin{figure*}[t]
%     \centering
%     \includegraphics[width=1\linewidth]{Figures/2.pdf}
%     \caption{Overall pipeline of BiSA. Dense visual tokens extracted by DINOv2 interact with learnable slots through iterative Bidirectional Slot Attention, and the refined slots are encoded into a compact global descriptor.}
%     \label{fig:method}
% \end{figure*}

\section{Related Work}

Visual retrieval matches corresponding images from large-scale databases using compact descriptors. We consider three tasks--VPR, CVGL, and ILIR--differing in appearance, geometry, and matching granularity but share the need to preserve stable structure and discriminative local evidence.

\noindent\textbf{Visual Place Recognition (VPR).}
VPR retrieves the same physical place under substantial appearance and viewpoint changes. Early aggregation heads match dense features against a fixed vocabulary of learned patterns, whether residual centroids in NetVLAD or mixing weights in MixVPR \cite{arandjelovic2016netvlad,alibey2023mixvpr}. With foundation-model backbones, the head no longer needs such a vocabulary: aggregation becomes a matter of solving how tokens are distributed over a small set of units, by optimal transport in SALAD, learnable queries in BoQ, or compact VLAD-style assignment in SuperVLAD \cite{izquierdo2024optimal,alibey2024boq,lu2024supervlad}. A parallel line instead intervenes outside the head, by selecting reliable regions (FoL-global), inserting retrieval tokens into the backbone (ImAge), or distilling into a deformable aggregator (D$^2$VPR) \cite{wang2025focus,lu2025image,zhang2026d2vpr}.

\noindent\textbf{Cross-View Geo-Localization (CVGL).}
CVGL matches geographic targets across ground, drone, and satellite viewpoints, where severe geometric variations make structural alignment important. Since SAFA's spatially aware aggregation \cite{shi2019spatial}, progress has come mainly from adding structural priors around the aggregation head: position-aware partitioning in CAMP, domain and scene alignment in DAC, multilevel enhancement in MEAN, and surface-aware guidance in SURFNet \cite{wu2024camp,xia2024dac,chen2025mean,liu2026surfnet}.

\noindent\textbf{Instance-Level Image Retrieval (ILIR).}
ILIR retrieves images of the same object or landmark, requiring compact descriptors to retain fine-grained local cues that distinguish visually similar instances. Aggregation here evolves from a single pooling rule, as in GeM, toward global--local or spatial attention-based weighting in SOLAR and GLAM \cite{radenovic2018fine,ng2020solar,song2022glam}. More recent designs add a second pathway for local evidence: DToP couples global and local branches over multi-layer features, and CiDeR localizes objects of interest within a single-stage pipeline \cite{song2023dtop,song2024cider}.

Across all three tasks, the aggregation head itself stays a one-shot mapping; structure and detail are coordinated, if at all, by modules placed around it. BiSA instead keeps both inside the head: one affinity, read along two axes under asymmetric temperatures, spans the granularity spectrum, while PTF uses the resulting axis-wise responses to down-weight unreliable tokens before the next iteration.

\section{Method}
\subsection{Overall Pipeline}

% As illustrated in Fig.~\ref{fig:method}, our framework reformulates dense-token aggregation as an iterative coupled process that produces a compact global
% descriptor. Given an image $I$, DINOv2~\cite{oquab2024dinov2} extracts $N=HW$ patch tokens of dimension $D$. Since stable structures and discriminative details are both
% defined by their spatial organization, a coordinate MLP embeds the 2D patch positions, yielding position-enhanced tokens
% $X_e\in\mathbb{R}^{N\times D}$.

Our framework reformulates dense-token aggregation as an iterative coupled process that produces a compact global
descriptor. Given an image $I$, DINOv2~\cite{oquab2024dinov2} extracts $N=HW$ patch tokens of dimension $D$. Since stable structures and discriminative details are both
defined by their spatial organization, a coordinate MLP embeds the 2D patch positions, yielding position-enhanced tokens
$X_e\in\mathbb{R}^{N\times D}$.

\begin{figure*}[t]
    \centering
    \includegraphics[width=1\linewidth]{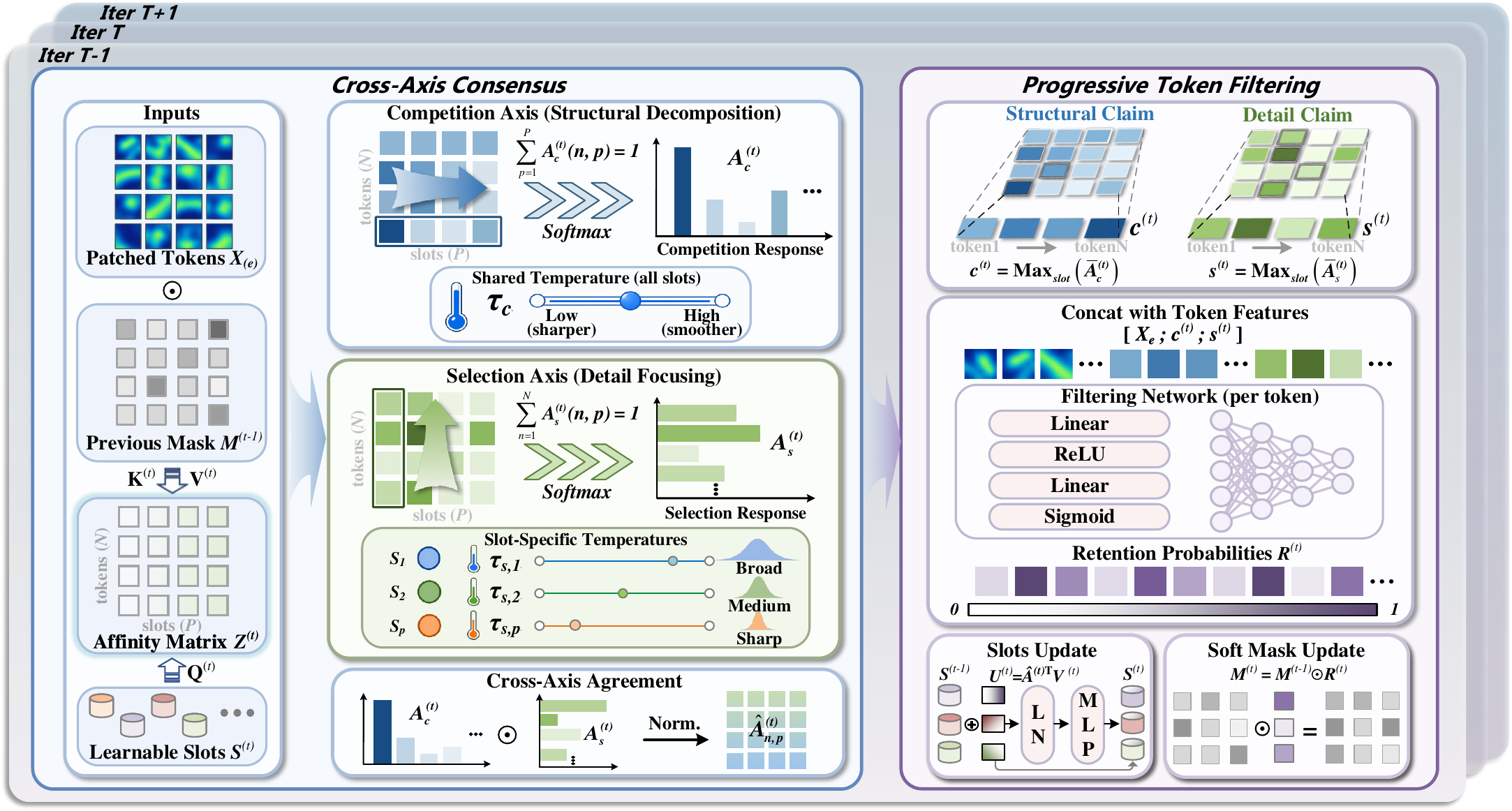}
    \caption{Illustration of a single BiSA iteration. CAC couples non-redundant scene decomposition with slot specialization across the structure--detail spectrum, while PTF updates the token-retention mask for the next iteration.}
    \label{fig:bisa}
\end{figure*}

The position-enhanced tokens are then fed into BiSA for iterative aggregation. Within BiSA, each iteration maintains a coupled state $(S^{(t)},M^{(t)})$, where $S^{(t)}\in\mathbb{R}^{P\times D}$ denotes evolving slots $P$ that aggregate local token evidence, and $M^{(t)}\in(0,1]^{N\times1}$ is a soft token-retention mask that controls each token's contribution to the next iteration. BiSA starts from $S^{(0)}=S_{\mathrm{init}}+\eta$, where
$S_{\mathrm{init}}\in\mathbb{R}^{P\times D}$ is learnable, shared across
images, and initialized from $\mathcal{N}(0,I)$, while
$\eta\sim\mathcal{N}(0,\sigma^2I)$ with $\sigma=0.05$ is added only during
training; $M^{(0)}=\mathbf{1}^{N\times1}$ allows all tokens to participate
equally in the first iteration. At each iteration, Cross-Axis Consensus (CAC) refines
the slots using token--slot associations jointly supported by competition and selection, while PTF updates the mask from their complementary reliability cues. The updated slots and reweighted tokens are jointly passed to the next iteration, allowing them to co-evolve.

After $T$ iterations of coupled refinement, the slots encode complementary yet interdependent scene evidence. A lightweight Transformer encoder~\cite{vaswani2017attention} therefore models their cross-slot dependencies before readout, giving
$\widetilde{S}=\operatorname{TransEnc}\!\left(S^{(T)}\right)\in\mathbb{R}^{P\times D}$.
Each refined slot is then linearly projected by
$W_o\in\mathbb{R}^{D\times D_o}$ and concatenated along the slot dimension, and the result is $\ell_2$-normalized to form the final descriptor $g\in\mathbb{R}^{PD_o}$, which is used for nearest-neighbor retrieval.

\subsection{Bidirectional Slot Attention}
\noindent\textbf{Cross-Axis Consensus (CAC).}
At iteration $t$, the previous slot state produces the queries $Q^{(t)}=\operatorname{LN}_{s}(S^{(t-1)})W_q\in\mathbb{R}^{P\times d}$, while the position-enhanced tokens produce keys and values that are jointly gated by the previous retention mask: $[K^{(t)},V^{(t)}] = M^{(t-1)}\odot \operatorname{LN}_{x}(X_e)[W_k,W_v] \in\mathbb{R}^{N\times2d}$. The shared token--slot affinity is computed as:
\begin{equation}
Z^{(t)} = \frac{ K^{(t)}{Q^{(t)}}^\top}{\sqrt d},\quad Z^{(t)}\in\mathbb{R}^{N\times P},
\label{eq:shared_affinity}
\end{equation}
where $d$ is the attention-head dimension. The competition axis first reads this shared affinity along the competition axis to determine which slot should own each token. Accordingly, $Z^{(t)}$ is normalized over the $P$ slots:
\begin{equation}
A_{\mathrm{c}}^{(t)}(n,p)
=
\frac{
\exp\!\left(Z_{n,p}^{(t)}/\tau_c\right)
}{
\sum_{p'=1}^{P}
\exp\!\left(Z_{n,p'}^{(t)}/\tau_c\right)
},
\
\sum_{p=1}^{P}
A_{\mathrm{c}}^{(t)}(n,p)=1,
\label{eq:competition_axis}
\end{equation}
where $\theta_c\in\mathbb{R}$ is a learnable scalar log-temperature, and $\tau_c=\operatorname{clip}(e^{\theta_c},\tau_{\min},\tau_{\max})
\in\mathbb{R}_{>0}$ is shared across all $P$ slots.
Each token distributes its probability mass among competing slots, while the shared temperature places all slots on a consistent competition scale,
enabling coherent decomposition into complementary scene components.

Competition alone determines the assignment of tokens, but does not control how broadly or sharply individual slots search for discriminative evidence. This distinction is crucial for retrieval, where evidence spans a structure--detail spectrum, requiring slot responses of varying support from broad layouts to sparse local cues. Accordingly, we read the same affinity
along the selection axis by normalizing it over tokens:
\begin{equation}
A_{\mathrm{s}}^{(t)}(n,p)
=
\frac{
\exp\!\left(Z_{n,p}^{(t)}/\tau_{s,p}\right)
}{
\sum_{n'=1}^{N}
\exp\!\left(Z_{n',p}^{(t)}/\tau_{s,p}\right)
},
\
\sum_{n=1}^{N}
A_{\mathrm{s}}^{(t)}(n,p)=1,
\label{eq:selection_axis}
\end{equation}
where $\boldsymbol{\theta}_s =(\theta_{s,1},\ldots,\theta_{s,P})\in\mathbb{R}^{P}$
is the vector of learnable slot-specific log-temperatures. Its $p$-th component determines
$\tau_{s,p} =\operatorname{clip}(e^{\theta_{s,p}},
\tau_{\min},\tau_{\max})$. Collectively, $\boldsymbol{\tau}_s =(\tau_{s,1},\ldots,\tau_{s,P})
\in\mathbb{R}_{>0}^{P}$, whereas $\tau_c\in\mathbb{R}_{>0}$ is shared across all slots. Small selection temperatures concentrate a slot on sparse localized cues, whereas large temperatures spread its response over broad spatial layouts. The learned slot-specific temperatures therefore organize the slots along a continuous structure--detail spectrum, from detail-focused to structure-oriented responses. To retain the competition-induced scene decomposition while incorporating these slot-specific response ranges, we multiply the two responses element-wise. After normalization over tokens, the consensus weights compute a weighted sum of $V^{(t)}$ for each slot:
\begin{equation}
\begin{gathered}
\widehat A_{n,p}^{(t)}
=
\frac{\bigl(A_{\mathrm{c}}^{(t)}\odot A_{\mathrm{s}}^{(t)}\bigr)_{n,p}+\epsilon}
{\sum_{n'=1}^{N}\left[\bigl(A_{\mathrm{c}}^{(t)}\odot A_{\mathrm{s}}^{(t)}\bigr)_{n',p}+\epsilon\right]},
\\[2pt]
U^{(t)}={\left(\widehat A^{(t)}\right)}^{\top} V^{(t)},
\
S^{(t)}=\operatorname{SlotUpdate}\!\left(S^{(t-1)},U^{(t)}\right).
\end{gathered}
\label{eq:consensus_aggregation}
\end{equation}
The element-wise product favors token--slot associations supported by both competition-based ownership and selection-based salience. Here, $U^{(t)}\in\mathbb{R}^{P\times D}$ is the consensus-pooled slot update, and $\operatorname{SlotUpdate}$ integrates it with $S^{(t-1)}$ through residual
addition, layer normalization, and a residual MLP, preserving prior slot content while incorporating new token information.

\begin{figure}
    \centering
    \includegraphics[width=1\linewidth]{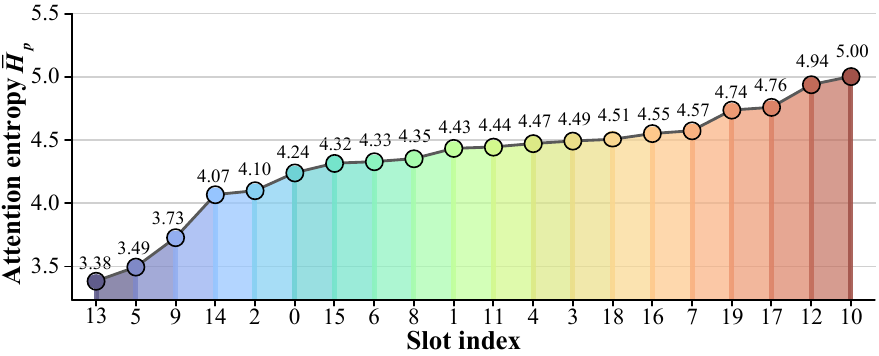}
    \caption{Per-slot response-entropy spectrum over 1{,}000 images.
    $\bar{H}_p$ is the entropy of slot $p$'s consensus response over the
    $N$ patch tokens (Eq.~\ref{eq:slot_entropy}). Slots are sorted by
    $\bar{H}_p$, from sharp slots isolating fine-grained details to broad
    slots covering global structure.}
    \label{fig:entropy}
\end{figure}

\noindent\textbf{Response spectrum analysis.}
To verify slot specialization along the structure--detail spectrum, we measure the entropy of each slot's final-iteration ($t\!=\!T$) consensus response:
\begin{equation}
\bar{H}_p = \frac{1}{|\mathcal{D}|}\sum_{I\in\mathcal{D}}
\Big(-\!\sum_{n=1}^{N}\widehat{A}^{(T)}_{n,p}(I)\,
\ln \widehat{A}^{(T)}_{n,p}(I)\Big),
\label{eq:slot_entropy}
\end{equation}
where $\mathcal{D}$ is the evaluation set and $I$ an image in it. Since $\widehat{A}^{(T)}_{\cdot,p}(I)$ is normalized over tokens (Eq.~\ref{eq:consensus_aggregation}), $\bar{H}_p$ is the response entropy of slot $p$ averaged across images. As shown in Fig.~\ref{fig:entropy}, the $P=20$ slots span a smooth entropy range of $\Delta H=\max_p\bar{H}_p-\min_p\bar{H}_p=1.62$: low-entropy slots concentrate on sparse localized cues, whereas high-entropy slots cover broad spatial layouts, revealing a continuous spectrum from detail-focused to structure-oriented responses.

\noindent\textbf{Progressive Token Filtering (PTF).} To allow subsequent slot competition, scene decomposition, and refinement to operate on an increasingly reliable token pool, we introduce PTF, which exploits claim strength and salience as
intrinsic reliability cues from competition and selection, respectively.
Specifically, PTF averages the two responses over attention heads to obtain
$\bar A_{\mathrm c}^{(t)}$ and $\bar A_{\mathrm s}^{(t)}$, takes their maxima
over slots as $c^{(t)},s^{(t)}\in\mathbb{R}^{N}$, and combines these scores
with the position-enhanced features $X_e$ to estimate per-token retention
probabilities and update the mask:
\begin{equation}
\begin{gathered}
c_n^{(t)}
= \max_{1\leq p\leq P}\bar A_{\mathrm c}^{(t)}(n,p), \
s_n^{(t)} = \max_{1\leq p\leq P}\bar A_{\mathrm s}^{(t)}(n,p),
\\[2pt]
R^{(t)}
= \sigma\!\left(
W_2\delta\!\left(W_1[X_e;c^{(t)};s^{(t)}]\right)\right)\in(0,1)^{N\times1},
\\[2pt]
M^{(t)} = M^{(t-1)}\odot R^{(t)},
\end{gathered}
\label{eq:progressive_filter}
\end{equation}
where $c_n^{(t)}$ measures how decisively token $n$ is assigned among competing slots, whereas $s_n^{(t)}$ measures how salient it is relative to other tokens within its most responsive slot. $\delta(\cdot)$ and $\sigma(\cdot)$ denote ReLU and sigmoid, respectively. PTF relies only on
token features and the two CAC-derived cues, without additional token-level or segmentation supervision.

The updated mask does not modify the already computed $S^{(t)}$. Instead, it
reweights the keys and values in iteration $t+1$. Consequently, CAC and PTF
jointly evolve the coupled state $(S^{(t)},M^{(t)})$, allowing subsequent
competition, decomposition, and refinement to operate on progressively
filtered visual evidence and form a self-reinforcing loop.

\subsection{Training Objective}
BiSA is optimized using Multi-Similarity loss
$\mathcal{L}_{\mathrm{MS}}$~\cite{wang2019multi}:
\begin{equation}
\begin{aligned}
\mathcal{L}_{\mathrm{MS}}
={}&
\frac{1}{B}
\sum_{i=1}^{B}
\frac{1}{\alpha}
\log
\left(
1+
\sum_{p\in\mathcal{P}_{i}}
e^{-\alpha(s_{ip}-\lambda)}
\right)
\\
&+
\frac{1}{B}
\sum_{i=1}^{B}
\frac{1}{\beta}
\log
\left(
1+
\sum_{n\in\mathcal{N}_{i}}
e^{\beta(s_{in}-\lambda)}
\right),
\end{aligned}
\label{eq:ms_loss}
\end{equation}
where $s_{ij}=g_i^\top g_j$, while $\mathcal{P}_i$ and
$\mathcal{N}_i$ denote the positive and negative sets for anchor $i$,
respectively. The model is trained solely with image-level retrieval
supervision; no additional supervision is introduced for slot decomposition
or PTF.

\section{Experiments}

We evaluate BiSA on three retrieval tasks: VPR, CVGL, and instance-level image retrieval. All three tasks use the same core BiSA aggregator, with 20 learnable slots, 3 iterations, and a 5120-dimensional global descriptor. Task-specific fine-tuning, objectives, and data protocols follow the conventions of each domain to ensure fair comparison with prior work.
\subsection{Benchmarks and Performance Evaluation}

\begin{table*}[t]
\centering
\tabcolsep=3.2pt
\renewcommand{\arraystretch}{0.9}
\renewcommand{\vn}[1]{{\scriptsize(#1)}}
\fitwidth{{\small
\begin{tabular}{l|c|c||cc||cc||cc||cc||cc||cc}
\toprule
\multirow{2}{*}{Method} &
\multirow{2}{*}{Dim} &
\multirow{2}{*}{Back.} &
\multicolumn{2}{c||}{Baidu Mall} &
\multicolumn{2}{c||}{SPED} &
\multicolumn{2}{c||}{Nordland$^{*}$} &
\multicolumn{2}{c||}{MSLS-val} &
\multicolumn{2}{c||}{Pitts30k} &
\multicolumn{2}{c}{SVOX-Overcast} \\
\cmidrule(lr){4-15}
& & &
R@1 & R@5 & R@1 & R@5 & R@1 & R@5 &
R@1 & R@5 & R@1 & R@5 & R@1 & R@5 \\
\midrule
NetVLAD \vn{CVPR'16} & 32768 & VGG16
& 53.1 & 70.5 & 70.2 & 84.5 & 32.6 & 47.1 & 53.1 & 66.5 & 81.9 & 91.2 & -- & -- \\
MixVPR$^{\ddagger}$ \vn{WACV'23} & 4096 & ResNet50
& 64.4 & 80.3 & 85.2 & 92.1 & 58.4 & 74.6 & 88.0 & 92.7 & 91.5 & 95.5 & 96.2 & 98.2 \\
SuperVLAD$^{\ddagger}$ \vn{NeurIPS'24} & 3072 & DINOv2-B
& 68.8 & 81.0 & 90.0 & 95.1 & 60.5 & 76.0 & 91.2 & 95.9 & 92.3 & 96.7 & 97.4 & 99.0 \\
SALAD \vn{CVPR'24} & 8448 & DINOv2-B
& 68.4 & 81.6 & \underline{92.1} & \underline{96.2} & 76.0 & 89.2 & 92.2 & 96.4 & 92.3 & 96.2 & 97.5 & 99.0 \\
BoQ \vn{CVPR'24} & 12288 & DINOv2-B
& 67.3 & 79.4 & 90.6 & 95.7 & \underline{80.7} & \underline{92.1} & 92.8 & 96.1 & 92.6 & 96.7 & \underline{97.7} & 99.0 \\
FoL-global \vn{AAAI'25} & 8448 & DINOv2-L
& 63.5 & 77.7 & 91.9 & \underline{96.2} & 74.1 & 88.8 & 92.8 & \underline{96.9} & \underline{93.6} & \underline{96.9} & 97.4 & \underline{99.2} \\
ImAge \vn{NeurIPS'25} & 6144 & DINOv2-B
& \underline{70.6} & \underline{83.8} & 91.6 & 95.6 & 77.2 & 88.8 & \underline{93.0} & \textbf{97.0} & \textbf{94.0} & \textbf{97.2} & 97.6 & 99.0 \\
\rowcolor{tabhl2}%
D$^2$VPR \vn{AAAI'26} & 4096 & DINOv2-S
& 64.1 & 78.4 & 90.9 & 96.0 & 70.1 & 83.6 & 91.6 & 96.1 & 94.9$^{\dagger}$ & 97.3$^{\dagger}$ & \underline{97.7} & 98.7 \\
\midrule
\rowcolor{tabhl}%
\textbf{Ours} & 5120 & DINOv2-B
& \textbf{74.8} & \textbf{86.0}
& \textbf{92.8} & \textbf{96.4}
& \textbf{84.8} & \textbf{93.8}
& \textbf{93.8} & 96.6
& 92.9 & 96.6
& \textbf{98.9} & \textbf{99.4} \\
\rowcolor{tabhl}%
\textbf{$\Delta$ vs.\ Best} & -- & --
& \textbf{\textcolor{tabred}{+4.2}} & \textbf{\textcolor{tabred}{+2.2}}
& \textbf{\textcolor{tabred}{+0.7}} & \textbf{\textcolor{tabred}{+0.2}}
& \textbf{\textcolor{tabred}{+4.1}} & \textbf{\textcolor{tabred}{+1.7}}
& \textbf{\textcolor{tabred}{+0.8}} & \textbf{\textcolor{tabred}{-0.4}}
& \textbf{\textcolor{tabred}{-1.1}} & \textbf{\textcolor{tabred}{-0.6}}
& \textbf{\textcolor{tabred}{+1.2}} & \textbf{\textcolor{tabred}{+0.2}} \\
\bottomrule
\end{tabular}
}}
\caption{Comparison with state-of-the-art methods on six VPR benchmarks
(R@1/R@5, \%). Best and second-best results are in \textbf{bold} and
\underline{underlined}. The $\Delta$ row reports the gap to the best
baseline, excluding D$^2$VPR on Pitts30k due to its cross-image encoder.
Results marked $\ddagger$ are re-evaluated by us under the same protocol;
the others are quoted from the original papers.}
\label{tab:vpr_results}
\end{table*}

\begin{table*}[t]
\centering
\tabcolsep=2.4pt
\renewcommand{\arraystretch}{1.05}
\renewcommand{\vn}[1]{{\scriptsize(#1)}}
\fitwidth{{\small
\begin{tabular}{l||cc|cc|cc|cc||cc|cc|cc|cc}
\toprule
\multirow{3}{*}{Method} &
\multicolumn{8}{c||}{Drone $\to$ Satellite} &
\multicolumn{8}{c}{Satellite $\to$ Drone} \\
\cmidrule(lr){2-9}\cmidrule(lr){10-17}
& \multicolumn{2}{c|}{150\,m} & \multicolumn{2}{c|}{200\,m}
& \multicolumn{2}{c|}{250\,m} & \multicolumn{2}{c||}{300\,m}
& \multicolumn{2}{c|}{150\,m} & \multicolumn{2}{c|}{200\,m}
& \multicolumn{2}{c|}{250\,m} & \multicolumn{2}{c}{300\,m} \\
\cmidrule(lr){2-17}
& R@1 & AP & R@1 & AP & R@1 & AP & R@1 & AP
& R@1 & AP & R@1 & AP & R@1 & AP & R@1 & AP \\
\midrule
MCCG \vn{TCSVT'24}
& 57.62 & 62.80 & 66.83 & 71.60 & 74.25 & 78.35 & 82.55 & 85.27
& 61.25 & 53.51 & 82.50 & 67.06 & 81.25 & 74.99 & 87.50 & 80.20 \\
S4G \vn{ICCV'23}
& 70.05 & 74.93 & 80.68 & 83.90 & 87.35 & 89.72 & 90.03 & 91.91
& 83.75 & 73.83 & 91.25 & 83.42 & 93.75 & 89.07 & 93.75 & 90.66 \\
CAMP \vn{TGRS'24}
& 78.90 & 82.38 & 86.83 & 89.28 & 91.95 & 93.63 & 95.68 & \underline{96.65}
& 87.50 & 78.98 & 95.00 & 87.05 & 95.00 & 91.05 & 96.25 & 93.44 \\
DAC \vn{TCSVT'24}
& 76.65 & 80.56 & 86.45 & 89.00 & 92.95 & 94.18 & 94.53 & 95.45
& 87.50 & 79.87 & \underline{96.25} & 88.98 & 95.00 & 92.81 & 96.25 & 94.00 \\
MEAN \vn{TGRS'25}
& 81.73 & 87.72 & 89.05 & 91.00 & 92.13 & 93.60 & 94.63 & 95.76
& \underline{91.25} & 81.50 & \underline{96.25} & 89.55 & 95.00 & 92.36 & 96.25 & 94.32 \\
\rowcolor{tabhl2}%
SURFNet \vn{TGRS'26}
& \underline{86.10} & \underline{88.22} & \underline{91.88} & \underline{93.20} & \underline{95.07} & \underline{95.83} & \underline{96.00} & 96.55
& \underline{91.25} & \underline{86.67} & \underline{96.25} & \underline{92.23} & \underline{97.50} & \underline{95.03} & \underline{98.75} & \underline{95.89} \\
\midrule
\rowcolor{tabhl}%
\textbf{Ours}
& \textbf{94.50} & \textbf{95.54} & \textbf{97.40} & \textbf{97.93}
& \textbf{98.68} & \textbf{98.98} & \textbf{99.40} & \textbf{99.54}
& \textbf{98.75} & \textbf{94.83} & \textbf{100.00} & \textbf{97.65}
& \textbf{100.00} & \textbf{98.47} & \textbf{100.00} & \textbf{98.78} \\
\rowcolor{tabhl}%
\textbf{$\Delta$ vs.\ Best}
& \textbf{\textcolor{tabred}{+8.40}} & \textbf{\textcolor{tabred}{+7.32}} & \textbf{\textcolor{tabred}{+5.52}} & \textbf{\textcolor{tabred}{+4.73}}
& \textbf{\textcolor{tabred}{+3.61}} & \textbf{\textcolor{tabred}{+3.15}} & \textbf{\textcolor{tabred}{+3.40}} & \textbf{\textcolor{tabred}{+2.89}}
& \textbf{\textcolor{tabred}{+7.50}} & \textbf{\textcolor{tabred}{+8.16}} & \textbf{\textcolor{tabred}{+3.75}} & \textbf{\textcolor{tabred}{+5.42}}
& \textbf{\textcolor{tabred}{+2.50}} & \textbf{\textcolor{tabred}{+3.44}} & \textbf{\textcolor{tabred}{+1.25}} & \textbf{\textcolor{tabred}{+2.89}} \\
\bottomrule
\end{tabular}
}}
\caption{Zero-shot generalization from University-1652 to SUES-200 using
R@1 and AP (\%). Best and second-best results are shown in \textbf{bold} and
\underline{underlined}. The $\Delta$ row reports the gap to the best baseline.}
\label{tab:cvgl_results}
\end{table*}

\begin{table}[t]
\centering
\tabcolsep=2.8pt
\renewcommand{\arraystretch}{1.0}
\renewcommand{\vn}[1]{{\tiny(#1)}}
\fitwidth{{\small
\begin{tabular}{l||c c||cc|cc}
\toprule
\multirow{2}{*}{Method} & \multirow{2}{*}{Train} & \multirow{2}{*}{Dim}
& \multicolumn{2}{c|}{Medium} & \multicolumn{2}{c}{Hard} \\
\cmidrule(lr){4-7}
& & & $\mathcal{R}$Oxf & $\mathcal{R}$Par & $\mathcal{R}$Oxf & $\mathcal{R}$Par \\
\midrule
GeM \vn{TPAMI'19}  & SfM-120k & 2048 & 64.7 & 77.2 & 38.5 & 56.3 \\
SOLAR \vn{ECCV'20} & SfM-120k & 2048 & 52.5 & 70.9 & 27.1 & 46.7 \\
GLAM \vn{WACV'22}  & SfM-120k & \phantom{0}512 & 66.2 & 77.5 & 39.5 & 54.3 \\
DToP \vn{WACV'23}  & SfM-120k & --   & \textbf{68.5} & \underline{83.1} & \underline{43.0} & \underline{65.8} \\
\rowcolor{tabhl2}%
CiDeR \vn{CVPR'24} & SfM-120k & 2048 & 67.3 & 79.4 & 42.4 & 57.5 \\
\midrule
\rowcolor{tabhl}%
\textbf{Ours (BiSA)} & SfM-120k & 5120
& \underline{67.7} & \textbf{83.9} & \textbf{47.5} & \textbf{67.7} \\
\rowcolor{tabhl}%
\textbf{$\Delta$ vs.\ DToP} & & &
\textbf{\textcolor{tabred}{-0.8}} & \textbf{\textcolor{tabred}{+0.8}} & \textbf{\textcolor{tabred}{+4.5}} & \textbf{\textcolor{tabred}{+1.9}} \\
\bottomrule
\end{tabular}
}}
\caption{Comparison with state-of-the-art single-stage global-descriptor methods on $\mathcal{R}$Oxf/$\mathcal{R}$Par under the Medium and Hard protocols (mAP).}
\label{tab:roxford_results}
\end{table}

\noindent\textbf{VPR.}
We evaluate on Baidu Mall, SPED, Nordland$^{*}$ (VPR-Bench $\pm1$ frame split), MSLS-val, Pitts30k, and
SVOX-Overcast, covering repetitive indoor scenes, seasonal and illumination
changes, large-scale urban environments, viewpoint variations, and domain
shifts
\cite{sun2017dataset,chen2017deep,sunderhauf2013are,warburg2020mapillary,
arandjelovic2016netvlad,berton2021adaptive}.
We report R@1 and R@5 using standard protocols~\cite{zaffar2021vprbench}:
a 25\,m threshold for street-view benchmarks, a tolerance of $\pm1$ frame
for Nordland$^{*}$, and the official ground truth for Baidu Mall.

\noindent\textbf{CVGL.}
We train on University-1652~\cite{zheng2020university} and transfer zero-shot
to SUES-200~\cite{zhu2023sues}, which covers 200 scenes with drone imagery at
150, 200, 250, and 300\,m plus paired satellite views; we use its 80-scene
test split without target-domain fine-tuning.
At each altitude, Drone$\rightarrow$Satellite matches 4{,}000 drone queries to
200 satellite images and Satellite$\rightarrow$Drone matches 80 satellite
queries to 10{,}000 drone images, with the 120 training scenes kept as
distractors.
We report R@1 and AP in both directions, counting same-scene views as positive.

\noindent\textbf{ILIR.}
We train on SfM-120k~\cite{radenovic2018fine}, whose 120K images carry
geometrically verified relations from reconstructed SfM models and also
supervise the whitening stage.
We evaluate on Revisited Oxford ($\mathcal{R}$Oxf) and Revisited Paris
($\mathcal{R}$Par)~\cite{radenovic2018revisiting}, which pair 70 queries with
4{,}993 and 6{,}322 database images and probe matching under viewpoint changes,
occlusion, cropping, and visually similar architectural distractors.
We report mAP under the Medium and Hard protocols without the R1M distractor
set: Medium counts Easy and Hard images as positives, Hard only the Hard ones.

\subsection{Implementation Details}

All experiments use DINOv2-B and fine-tune only its final three Transformer blocks, training on a single NVIDIA RTX A6000 GPU.

\noindent\textbf{VPR.}
We train on GSV-Cities~\cite{alibey2022gsvcities} for 40 epochs using
Multi-Similarity loss~\cite{wang2019multi}, online hard mining, AdamW
\cite{loshchilov2019decoupled}, and batches of 200 places with four images
per place. The learning rates are $4\times10^{-4}$ for the aggregator and $1\times10^{-4}$ for the backbone, with train/test resolutions of $280\times280$/$322\times322$.

\noindent\textbf{CVGL.}
We train on University-1652~\cite{zheng2020university} for 80 epochs using the
same loss as in VPR, sampling 32 buildings with four images each. AdamW uses
learning rates of $2\times10^{-4}$ for BiSA and $5\times10^{-5}$ for the
backbone.

\noindent\textbf{ILIR.}
We train on SfM-120k for 15 epochs using a siamese contrastive loss with a
margin of 0.85~\cite{hadsell2006dimensionality}, epoch-wise global
hard-negative re-mining, and Adam with a learning rate of $10^{-6}$.
Images are resized such that the longer side is 512 while preserving the
aspect ratio; inference uses single-scale descriptors and supervised
whitening learned on SfM-120k.

\subsection{Comparisons with State-of-the-art Methods}

\noindent\textbf{VPR (Table~\ref{tab:vpr_results}).}
We compare BiSA with NetVLAD~\cite{arandjelovic2016netvlad},
MixVPR~\cite{alibey2023mixvpr},
SuperVLAD~\cite{lu2024supervlad},
SALAD~\cite{izquierdo2024optimal},
BoQ~\cite{alibey2024boq},
FoL-global~\cite{wang2025focus},
ImAge~\cite{lu2025image}, and
D$^2$VPR~\cite{zhang2026d2vpr}.
BiSA sets a new state-of-the-art in R@1 on five of the six benchmarks,
while using a substantially more compact 5120-dimensional descriptor
than SALAD and BoQ. Its advantage is most pronounced under severe appearance
variation and subtle discriminative cues: R@1 improves over the second-best
method by $+4.2$, $+4.1$, $+0.7$, and $+1.2$ on Baidu Mall, Nordland$^{*}$,
SPED, and SVOX-Overcast, respectively. The large gains on Baidu Mall and
Nordland$^{*}$ support our claim that BiSA coordinates detail-level discriminability and structural invariance across the structure--detail spectrum. 
On the largely saturated Pitts30k, BiSA remains competitive, showing that
these gains do not come at the expense of conventional scenes.

\noindent\textbf{CVGL (Table~\ref{tab:cvgl_results}).}
We compare BiSA with MCCG~\cite{shen2024mccg},
Sample4Geo~\cite{deuser2023sample4geo},
CAMP~\cite{wu2024camp},
DAC~\cite{xia2024dac},
MEAN~\cite{chen2025mean}, and
SURFNet~\cite{liu2026surfnet}.
Under zero-shot transfer from University-1652 to SUES-200, BiSA outperforms
all competing methods on all 16 metrics across both retrieval directions and
four altitudes. At 150\,m in Drone$\rightarrow$Satellite retrieval, it exceeds
the second-best method by $+8.40$ R@1 and $+7.32$ AP, while attaining
100\% R@1 at several altitudes in the Satellite$\rightarrow$Drone direction.
Given the extreme viewpoint, scale, and appearance disparities, these gains
without target-domain fine-tuning support the role of the competition axis in
organizing complementary structural cues and strengthening cross-domain
structural invariance.

\noindent\textbf{ILIR (Table~\ref{tab:roxford_results}).}
We compare BiSA with the single-stage global-descriptor methods
GeM~\cite{radenovic2018fine},
SOLAR~\cite{ng2020solar},
GLAM~\cite{song2022glam},
DToP~\cite{song2023dtop}, and
CiDeR~\cite{song2024cider}.
BiSA outperforms CiDeR on all four $\mathcal{R}$Oxf/$\mathcal{R}$Par Medium and Hard metrics. Compared with the stronger ViT-hybrid DToP, BiSA leads on three of the four metrics, including gains of $+4.5$ on $\mathcal{R}$Oxf-Hard and $+1.9$ on $\mathcal{R}$Par-Hard. Since the Hard protocol most strongly tests fine-grained discrimination
against confusing distractors, these margins support the role of the selection
axis in focusing on sparse discriminative local evidence and strengthening
detail-level discriminability. BiSA trails DToP only
on $\mathcal{R}$Oxf-Medium by 0.8 points. It achieves this overall performance with a simpler pure-ViT backbone and a standard siamese contrastive loss, indicating
that the gains arise primarily from the BiSA aggregation mechanism.

\begin{figure}
    \centering
    \includegraphics[width=1\linewidth]{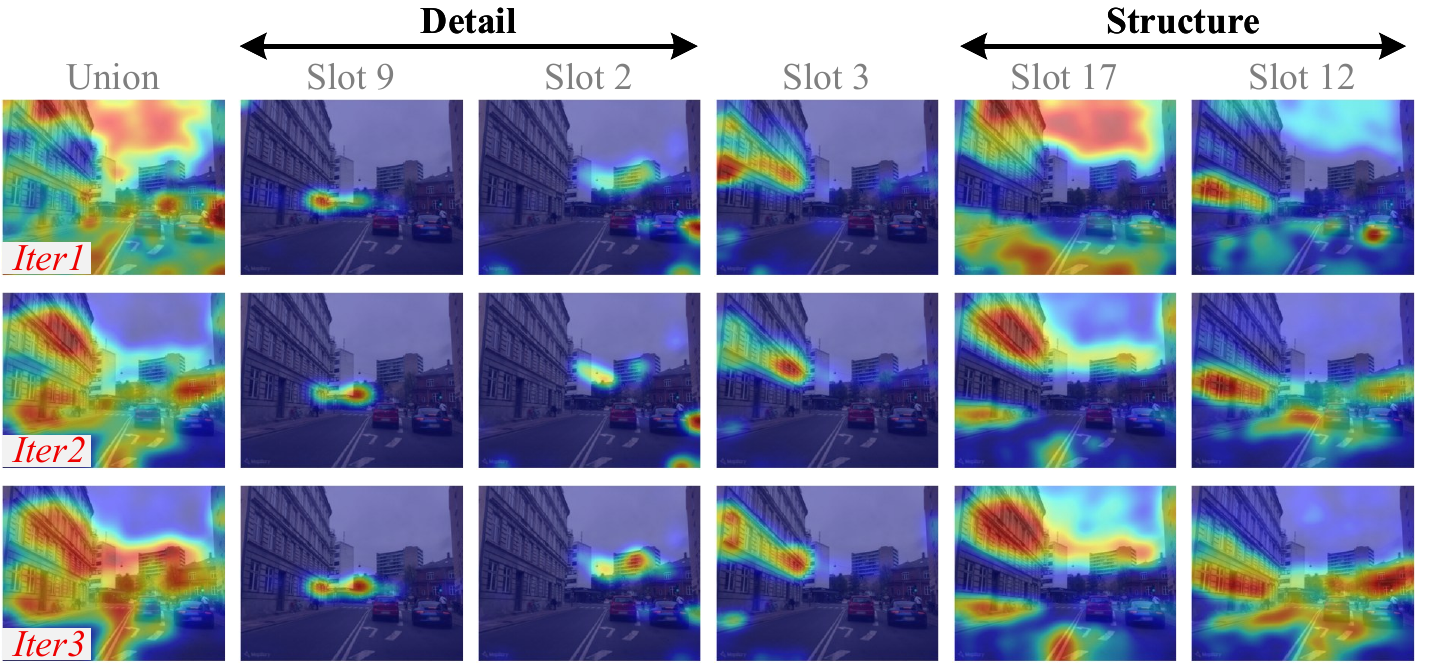}
    \caption{Consensus heatmaps of BiSA across iterations, showing a continuum from detail-focused low-entropy slots to structure-focused high-entropy slots.}
    \label{fig:heatmaps}
\end{figure}

Figure~\ref{fig:heatmaps} visualizes representative BiSA consensus heatmaps
across iterations. Consistent with the entropy analysis in
Figure~\ref{fig:entropy}, the slots span a response spectrum from sparse
fine-grained details to broad scene layouts. Low-entropy slots (e.g., Slots 9
and 2) sharply localize discriminative cues, whereas high-entropy slots
(e.g., Slots 17 and 12) cover broader structures such as roads and building
contours. Their responses become progressively more coherent across
iterations, qualitatively supporting BiSA's multi-granular slot
specialization and iterative structure--detail consensus.

\begin{figure}
    \centering
    \includegraphics[width=1\linewidth]{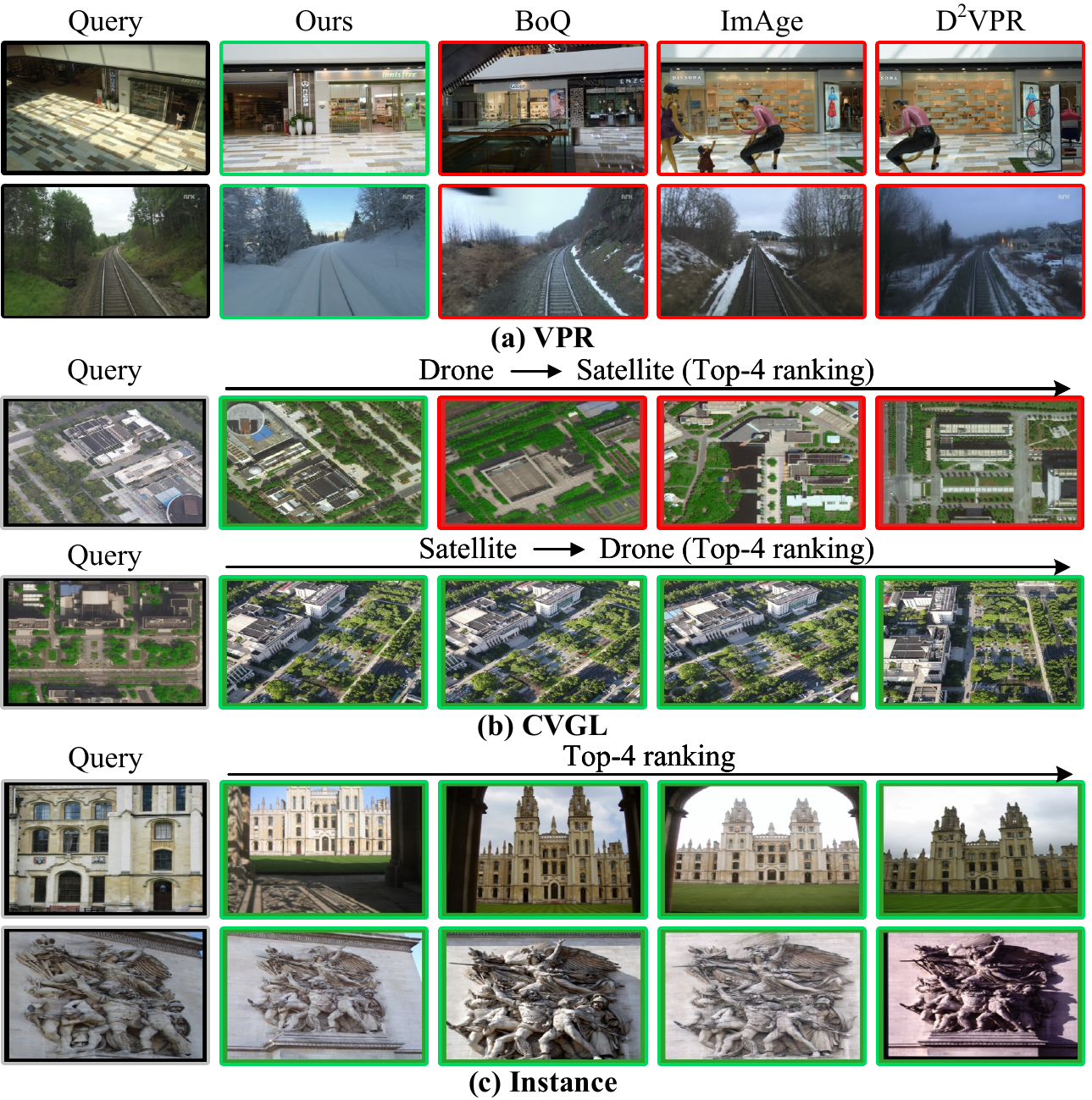}
    \caption{Qualitative results. (a) VPR: Top-1 retrievals of BiSA and prior methods (BoQ, ImAge, D$^2$VPR); (b) CVGL: Top-4 results for bidirectional Drone$\leftrightarrow$Satellite retrieval; (c) Instance retrieval: Top-4 rankings on $\mathcal{R}$Oxford and $\mathcal{R}$Paris. Green/red outlines mark correct/incorrect retrievals.}
    \label{fig:Visual}
\end{figure}

Figure~\ref{fig:Visual} compares BiSA with representative methods across the
three retrieval tasks. In VPR, BiSA retrieves the correct place under perceptual aliasing and severe appearance changes. In CVGL, it ranks the target highly despite substantial viewpoint and scale gaps. In ILIR, it promotes correct landmarks over confusing distractors. These results support the competition axis in organizing complementary scene content and the
selection axis in spanning broad layouts to sparse discriminative cues, allowing BiSA to reconcile structural invariance with detail-level discriminability.

\subsection{Ablation Study}

\begin{table}[t]
\centering
\tabcolsep=2.8pt
\renewcommand{\arraystretch}{1}
\fitwidth{{\small
\begin{tabular}{l||c|c|c||c}
\toprule
Method & Baidu M. & Nordland$^{*}$ & MSLS & Avg. \\
\midrule
\multicolumn{5}{c}{\textbf{\textit{Cross-Axis Consensus}}} \\
\midrule
No-competition baseline
& 68.9 & 78.8 & 88.9 & 78.9 \\
Competition-only baseline
& 69.4 & 79.3 & 89.6 & 79.4 \\
+ Selection axis
& 72.6 & 82.2 & 91.8 & 82.2 \\
+ Asymmetric temperature
& 73.1 & 83.1 & 92.2 & 82.8 \\
\midrule
\multicolumn{5}{c}{\textbf{\textit{Iterative Coupling}}} \\
\midrule
Baseline
& 73.1 & 83.1 & 92.2 & 82.8 \\
+ Independent mask
& 74.3 & 84.2 & 93.6 & 84.0 \\
\midrule
\rowcolor{tabhl}%
\textbf{Full BiSA (PTF)}
& \textbf{74.8} & \textbf{84.8} & \textbf{93.8} & \textbf{84.5} \\
\rowcolor{tabhl}%
\textbf{$\Delta$ vs.\ No-comp.}
& \textbf{\textcolor{tabred}{+5.9}} & \textbf{\textcolor{tabred}{+6.0}} & \textbf{\textcolor{tabred}{+4.9}} & \textbf{\textcolor{tabred}{+5.6}} \\
\bottomrule
\end{tabular}
}}
\caption{Ablation of BiSA on VPR benchmarks (R@1, \%). The $\Delta$ row reports the improvement of Full BiSA over the no-competition baseline. All variants share the same backbone, training protocol, and 5120-dimensional descriptor.}
\label{tab:ablation}
\end{table}

% We conduct all ablations in the VPR setting using Baidu Mall, Nordland, and
% MSLS-val, which respectively emphasize fine-grained discrimination, seasonal
% structural robustness, and large-scale place recognition. In the
% iterative-coupling block, the baseline uses full CAC without filtering. The
% independent-mask variant applies a newly predicted retention mask at each
% iteration, whereas PTF progressively accumulates these masks across iterations
% to form Full BiSA.

We conduct all ablations in the VPR setting on Baidu Mall, Nordland$^{*}$, and
MSLS-val, which emphasize fine-grained discrimination, seasonal structural
robustness, and large-scale place recognition, respectively. In the
iterative-coupling block, the baseline uses full CAC without filtering; the
independent-mask variant applies a newly predicted retention mask at each
iteration, whereas PTF accumulates these masks across iterations to form
Full BiSA. We further analyze the effect of descriptor dimensionality and
the associated retrieval cost.

% \noindent\textbf{Cross-Axis Consensus.}
% Starting from the no-competition baseline, token-to-slot competition improves
% R@1 by 0.5, 0.5, and 0.7 points on Baidu Mall, Nordland, and MSLS-val. In the same order, the selection axis with only 20 learnable scalars adds 3.2, 2.9, and 2.2 points,
% averaging 2.8, and asymmetric temperature control a further 0.5, 0.9,
% and 0.4 by adapting slot-wise support breadth. Competition alone yields modest gains because relative token ownership does
% not control each slot's support breadth.
% Together, the CAC components raise average R@1 from 78.9 to 82.8.

% \noindent\textbf{Iterative Coupling.}
% With full CAC fixed, independently applying the retention mask predicted at
% each iteration improves R@1 by 1.2, 1.1, and 1.4 points on Baidu Mall,
% Nordland, and MSLS-val, respectively, increasing the average from 82.8 to
% 84.0. PTF progressively accumulates these masks across iterations, further
% improving R@1 by 0.5, 0.6, and 0.2 points, respectively, and raising the
% average to 84.5. These results confirm that CAC and PTF are complementary:
% CAC couples non-redundant decomposition with multi-granular slot
% specialization, while PTF enables subsequent iterations to operate on
% increasingly reliable token evidence.

\noindent\textbf{Cross-Axis Consensus.}
Starting from the no-competition baseline, token-to-slot competition improves
R@1 by 0.5, 0.5, and 0.7 points on Baidu Mall, Nordland$^{*}$, and MSLS-val. In the
same order, the selection axis with only 20 learnable scalars adds 3.2, 2.9,
and 2.2 points, averaging 2.8, which is the largest single gain in
Table~\ref{tab:ablation} and accounts for half of the total improvement of
Full BiSA. Asymmetric temperature control adds a further 0.5, 0.9, and 0.4
points by adapting slot-wise support breadth, with the largest gain on
Nordland. Competition alone yields modest gains because relative token
ownership does not control each slot's support breadth; once the selection
axis supplies this control, the two axes become effective jointly. Together,
the CAC components raise average R@1 from 78.9 to 82.8, contributing about
70\% of the full improvement, and each component improves all three
benchmarks.

\noindent\textbf{Iterative Coupling.}
With full CAC fixed, independently applying the retention mask predicted at
each iteration improves R@1 by 1.2, 1.1, and 1.4 points on Baidu Mall,
Nordland$^{*}$, and MSLS-val, respectively, increasing the average from 82.8 to
84.0, with the largest gain on MSLS-val. PTF progressively accumulates these
masks across iterations, further improving R@1 by 0.5, 0.6, and 0.2 points,
respectively, and raising the average to 84.5. Token filtering thus
contributes the remaining 1.7 points, about 30\% of the full improvement, and
the additional gain of accumulation over independent masks shows that
retention decisions benefit from evidence gathered in earlier iterations.

Overall, Full BiSA improves R@1 by 5.9, 6.0, and 4.9 points over the
no-competition baseline. These results confirm that CAC and PTF are
complementary: CAC couples non-redundant decomposition with multi-granular
slot specialization, while PTF enables subsequent iterations to operate on
increasingly reliable token evidence.

% \begin{table}[t]
% \centering
% \tabcolsep=4pt
% \renewcommand{\arraystretch}{1.0}
% \fitwidth{{\small
% \begin{tabular}{l||c|c||c}
% \toprule
% Method & Dim & Index (GB) & Avg. R@1 \\
% \midrule
% MixVPR     & 4096  & 16.4 & 80.6 \\
% SuperVLAD  & 3072  & 12.3 & 83.4 \\
% D$^2$VPR   & 4096  & 16.4 & 84.9 \\
% FoL-global & 8448  & 33.8 & 85.6 \\
% SALAD      & 8448  & 33.8 & 86.4 \\
% BoQ        & 12288 & 49.2 & 87.0 \\
% ImAge      & 6144  & 24.6 & 87.3 \\
% \midrule
% \rowcolor{tabhl}%
% \textbf{Ours} & 5120 & 20.5 & \textbf{89.7} \\
% \bottomrule
% \end{tabular}
% }}
% \caption{Descriptor dimensionality, index size, and accuracy. Index size is
% the float32 storage for one million database images, to which exhaustive
% search cost is proportional. Avg.\ R@1 is averaged over the six benchmarks
% in Table~\ref{tab:vpr_results}.}
% \label{tab:dim_cost}
% \end{table}

\begin{table}[t]
\centering
\tabcolsep=3.5pt
\renewcommand{\arraystretch}{1.0}
\fitwidth{{\small
\begin{tabular}{l||c|c|c||c}
\toprule
Method & Dim$\downarrow$ & Storage (KB)$\downarrow$ & Lat. (ms)$\downarrow$ & Avg. R@1$\uparrow$ \\
\midrule
SuperVLAD  & \textbf{3072}  & \textbf{12.3} & \textbf{9.8}  & 80.6 \\
D$^2$VPR   & \underline{4096} & \underline{16.4} & 9.6$^{\dagger}$ & 82.3 \\
FoL-global & 8448  & 33.8 & 19.7 & 83.2 \\
SALAD      & 8448  & 33.8 & \underline{10.0} & 84.2 \\
BoQ        & 12288 & 49.2 & 11.1 & 84.8 \\
ImAge      & 6144  & 24.6 & 10.7 & \underline{85.3} \\
\midrule
\rowcolor{tabhl}%
\textbf{Ours} & 5120 & 20.5 & 11.9 & \textbf{87.8} \\
\bottomrule
\end{tabular}
}}
\caption{Descriptor dimensionality, retrieval cost, and accuracy. Best and
second-best results are shown in \textbf{bold} and \underline{underlined}.
Storage is the float32 size of one database descriptor; database storage and
exhaustive search cost scale linearly with it. Latency is the per-image
encoding time on an RTX 3080\,Ti (batch size 1, $322\times322$, mixed
precision). Avg.\ R@1 is averaged over Baidu Mall, SPED, Nordland$^{*}$, MSLS-val, and
Pitts30k. $^{\dagger}$Excludes the cross-image encoder of D$^2$VPR, which
is used for its Pitts30k result; not ranked.}
\label{tab:dim_cost}
\end{table}

\noindent\textbf{Descriptor dimensionality and retrieval cost.}
We examine whether BiSA's gains stem from aggregation or descriptor
capacity (Table~\ref{tab:dim_cost}). Accuracy does not follow
dimensionality: among DINOv2-B methods, the two largest descriptors, BoQ and
SALAD, trail the 6144-dimensional ImAge in average R@1, while BiSA leads
although its strongest competitors use 1.2--2.4$\times$ larger descriptors.
Moreover, all ablation variants share the same 5120-dimensional descriptor,
which falls below the best baselines on all three benchmarks without
competition and surpasses them with the BiSA components. The gains thus stem
from token organization rather than descriptor size.
Practically~\cite{lu2024cricavpr}, each BiSA descriptor occupies 20.5\,KB,
42\%, 61\%, and 83\% of those of BoQ, SALAD, and ImAge, reducing storage and
exhaustive search cost by the same ratios. Its iterative refinement adds
only 0.8\,ms over BoQ, whereas FoL-global scales to DINOv2-L, requiring
19.7\,ms while trailing by 4.6 points in average R@1.

\section{Conclusion}

In this work, we presented Bidirectional Slot Attention (BiSA), an iterative aggregation framework that reconciles structural invariance and detail-level discriminability across a continuous granularity spectrum without external semantic or structural priors. Cross-Axis Consensus (CAC) applies shared-temperature competition and slot-specific-temperature selection to a common token--slot affinity, coupling non-redundant scene decomposition with slot specialization from broad layouts to sparse discriminative details. Progressive Token Filtering (PTF) uses the resulting claim-strength and salience cues to suppress unreliable tokens and feed a refined token pool into subsequent iterations, forming a feedback loop of decomposition, focusing, and purification. Across VPR, CVGL, and ILIR, BiSA consistently improves retrieval with a compact descriptor, while zero-shot cross-view transfer without target-domain fine-tuning further supports iterative structure--detail consensus as a task-general aggregation principle.

\section*{LLM Usage Statement}

We used OpenAI ChatGPT to assist with language editing and figure refinement during manuscript preparation. Specifically, we used ChatGPT to edit and refine author-written text and author-produced figures. The assistance was limited to refining content produced by the authors and did not cover related work, code, experiments, data, or research ideas and conclusions. The authors reviewed all resulting text and figures against the original content and take responsibility for the final manuscript and its scientific claims.

{
    \small
    \bibliographystyle{ieeenat_fullname}
    \bibliography{main}
}

% Appendix content, migrated from the AAAI supplementary.tex.
% Compiled on its own by supplementary.tex; can also be appended to the
% main paper by uncommenting \input{sec/X_suppl} in main.tex.

\clearpage
\setcounter{page}{1}
\maketitlesupplementary

\appendix

\renewcommand{\thetable}{S\arabic{table}}
\renewcommand{\thefigure}{S\arabic{figure}}
\renewcommand{\theequation}{S\arabic{equation}}
\setcounter{table}{0}
\setcounter{figure}{0}
\setcounter{equation}{0}

\section{Benchmark Details and Evaluation Protocols}
\label{app:benchmarks}

We provide additional details on the datasets and evaluation protocols used
for visual place recognition (VPR), cross-view geo-localization (CVGL), and
instance-level image retrieval (ILIR). Table~\ref{tab:supp_benchmarks}
summarizes the dataset scales and positive-match criteria.

\subsection{Visual Place Recognition}

\noindent\textbf{Training data.}
GSV-Cities~\cite{alibey2022gsvcities} contains approximately 530K
street-view images from more than 62K places across over 40 cities and a
14-year period. Its geographic, temporal, and viewpoint diversity provides
broad training coverage for VPR.

\noindent\textbf{Evaluation benchmarks.}
Baidu Mall~\cite{sun2017dataset} contains repetitive indoor corridors and
storefronts, making retrieval dependent on subtle local cues under severe
perceptual aliasing. SPED~\cite{chen2017deep} consists of paired surveillance
views captured under substantial illumination, weather, and seasonal changes.
Nordland$^{*}$~\cite{sunderhauf2013are} provides time-aligned recordings of
the same railway route across four seasons. Following the VPR-Bench
setup~\cite{zaffar2021vprbench}, we use 2,760 summer queries against the
full 27,592-image winter reference sequence. This split, denoted
Nordland$^{*}$ throughout, differs from the 27,592-query split evaluated
within $\pm10$ frames in~\cite{alibey2024boq,lu2025image}; results on the
two splits are not comparable.
MSLS-val~\cite{warburg2020mapillary} covers diverse urban environments,
viewpoints, weather, illumination, and dynamic conditions. Pitts30k
~\cite{arandjelovic2016netvlad} primarily evaluates urban retrieval under
viewpoint variation, while SVOX-Overcast~\cite{berton2021adaptive} introduces
cross-platform and appearance shifts between Oxford RobotCar queries and a
Google Street View gallery.

\noindent\textbf{Protocol.}
We report R@1 and R@5 following standard VPR protocols
~\cite{zaffar2021vprbench}. A retrieval is considered positive within 25\,m
for MSLS-val, Pitts30k, and SVOX-Overcast, within $\pm1$ frame for Nordland$^{*}$,
and according to the official correspondences for Baidu Mall and SPED.

\subsection{Cross-View Geo-Localization}

\noindent\textbf{Training data.}
University-1652~\cite{zheng2020university} contains drone, satellite, and
ground views of 1,652 buildings from 72 universities. Its official training
split includes 50,218 images from 701 buildings across 33 universities. We
train BiSA using the drone and satellite views.

\noindent\textbf{Evaluation benchmark.}
SUES-200~\cite{zhu2023sues} contains 200 scenes with drone images captured
at altitudes of 150, 200, 250, and 300\,m, together with corresponding
satellite images; each scene provides 50 drone images per altitude and one
satellite image. We train exclusively on University-1652 and perform
zero-shot evaluation on the 80-scene SUES-200 test split without
target-domain fine-tuning.

\noindent\textbf{Protocol.}
At each altitude, Drone$\to$Satellite retrieval uses 4,000 drone queries and
200 satellite gallery images, whereas Satellite$\to$Drone retrieval uses
80 satellite queries and 10,000 drone gallery images. Following the official
protocol, the galleries contain images from all 200 scenes, with the 120
training scenes serving as additional distractors. We report R@1 and AP
for both directions, treating images of the same scene as positive matches.

\begin{table}[t]
\centering
\setlength{\tabcolsep}{3.2pt}
\renewcommand{\arraystretch}{1.08}
\fitwidth{{\small
\begin{tabular}{
@{}p{0.29\columnwidth}
p{0.63\columnwidth}@{}}
\toprule
\textbf{Dataset} & \textbf{Scale / Protocol} \\
\midrule

\multicolumn{2}{c}{\textbf{Visual Place Recognition}} \\
\midrule
GSV-Cities
& Train: 530K images / 62K places \\

Baidu Mall
& 2,292 / 689 Q--G; official annotations \\

SPED
& 607 / 607 Q--G; paired correspondence \\

Nordland$^{*}$
& 2,760 / 27,592 Q--G; within $\pm1$ frame \\

MSLS-val
& 740 / 18,871 Q--G; within 25\,m \\

Pitts30k-test
& 6,816 / 10,000 Q--G; within 25\,m \\

SVOX-Overcast
& 872 / 17,166 Q--G; within 25\,m \\

\midrule
\multicolumn{2}{c}{\textbf{Cross-View Geo-Localization}} \\
\midrule
University-1652
& Train: 50,218 images / 701 buildings \\

SUES D$\to$S
& 4,000 / 200 Q--G per altitude; same scene \\

SUES S$\to$D
& 80 / 10,000 Q--G per altitude; same scene \\

\midrule
\multicolumn{2}{c}{\textbf{Instance-Level Image Retrieval}} \\
\midrule
SfM-120k
& Train: approximately 120K images \\

$\mathcal{R}$Oxf
& 70 / 4,993 Q--G; Medium and Hard \\

$\mathcal{R}$Par
& 70 / 6,322 Q--G; Medium and Hard \\
\bottomrule
\end{tabular}
}}

\caption{Dataset statistics and evaluation protocols. Q--G denotes the
numbers of query and gallery images. SUES-200 statistics are reported
separately for each drone altitude.}
\label{tab:supp_benchmarks}
\end{table}

\subsection{Instance-Level Image Retrieval}

\noindent\textbf{Training data.}
SfM-120k~\cite{radenovic2018fine} contains approximately 120K landmark
images organized using automatically reconstructed Structure-from-Motion
models. The resulting geometrically verified image relations are used for
fine-tuning and supervised whitening.

\noindent\textbf{Evaluation benchmarks.}
Revisited Oxford ($\mathcal{R}$Oxf) and Revisited Paris
($\mathcal{R}$Par)~\cite{radenovic2018revisiting} each contain 70 landmark
queries and 4,993 or 6,322 database images, respectively. They evaluate
instance-level matching under viewpoint changes, occlusion, cropping, and
visually similar architectural distractors.

\noindent\textbf{Protocol.}
We report mAP under the Medium and Hard protocols without the R1M distractor
set. Under Medium, Easy and Hard images are treated as positives, while
Unclear images are ignored. Under Hard, only Hard images are treated as
positives, while Easy and Unclear images are ignored. Evaluation uses
single-scale descriptors and supervised whitening learned exclusively on
SfM-120k.

\section{Additional Training and Implementation Details}
\label{app:implementation}

\subsection{Common BiSA Configuration}

Unless otherwise noted, all tasks use DINOv2-B~\cite{oquab2024dinov2}, with
only its final three Transformer blocks fine-tuned, followed by a $3\times3$
convolution that projects features to 512-dimensional tokens. BiSA uses
$P=20$ learnable slots~\cite{locatello2020slot} and $T=3$ coupled iterations
with eight 64-dimensional attention heads and Q/K/V projections shared across
iterations. The competition axis uses one shared learnable temperature,
whereas the selection axis uses slot-specific temperatures; all are initialized
to 0.5 and constrained to $[0.05,1.0]$. The positional embedding is produced
by a $2\!\to\!512\!\to\!512$ MLP applied to normalized 2D coordinates and
added to the visual tokens. PTF predicts per-token retention
probabilities through a $514\!\to\!128\!\to\!1$ MLP using the token feature,
claim strength under competition, and salience under selection, with masks
accumulated multiplicatively across iterations. After $T$ iterations, a
one-layer, four-head Transformer encoder~\cite{vaswani2017attention} models
inter-slot interactions; each slot is projected to 256 dimensions, producing
a 5120-dimensional descriptor after concatenation. All models use mixed
precision and gradient clipping at 1.0. Unless otherwise specified, results
are averaged over three independent runs, with checkpoint selection and
evaluation performed separately for each run. VPR and CVGL use
Multi-Similarity loss with $\alpha=1$, $\beta=50$, and base value 0.0,
together with a Multi-Similarity Miner using $\epsilon=0.1$.

\subsection{Task-Specific Training Settings}

\noindent\textbf{VPR.}
We train on GSV-Cities for 40 epochs with a batch size of 200 places,
sampling four images per place, so that each batch contains 800 images in
total. Training and evaluation resolutions are $280\times280$ and
$322\times322$, respectively.
Training images are augmented using rand-augment with three operations
and magnitude 15. We use AdamW with learning rates of $4\times10^{-4}$ for
BiSA and $1\times10^{-4}$ for the backbone, weight decay
$5\times10^{-4}$, a five-epoch linear warm-up, and MultiStepLR milestones at
epochs 10, 20, and 30.

\noindent\textbf{CVGL.}
We train on University-1652 for 80 epochs, sampling 32 buildings with one
satellite and three drone images per building. Training and evaluation
resolutions are $224\times224$ and $252\times252$, respectively. Drone images
are augmented using random rotation, random resized cropping, and
rand-augment, while satellite images use random rotation, horizontal
flipping, and color jitter. We use AdamW with learning rates of
$2\times10^{-4}$ for BiSA and $5\times10^{-5}$ for the backbone, a three-epoch
warm-up, and learning-rate milestones at epochs 40, 60, and 70. Checkpoints
are selected according to University-1652 Drone$\to$Satellite R@1.

\noindent\textbf{ILIR.} We fine-tune on SfM-120k  for 15 epochs with a siamese contrastive loss~\cite{hadsell2006dimensionality} (margin 0.85) and Adam (learning rate $10^{-6}$, weight decay $10^{-6}$, exponential decay with $\gamma = 0.99$). Each epoch draws 2,000 query--positive pairs and re-mines five hard negatives per query from a random pool of 20,000 images, taking at most one negative per SfM cluster; five tuples form one optimizer step. Evaluation uses single-scale descriptors with supervised whitening learned from 8,000 SfM-120k pairs, applied without dimensionality reduction.

\section{Additional Qualitative Results}
\label{app:qualitative}

\subsection{Extended Retrieval Comparisons}

\begin{figure*}
    \centering
    \includegraphics[width=1\linewidth]{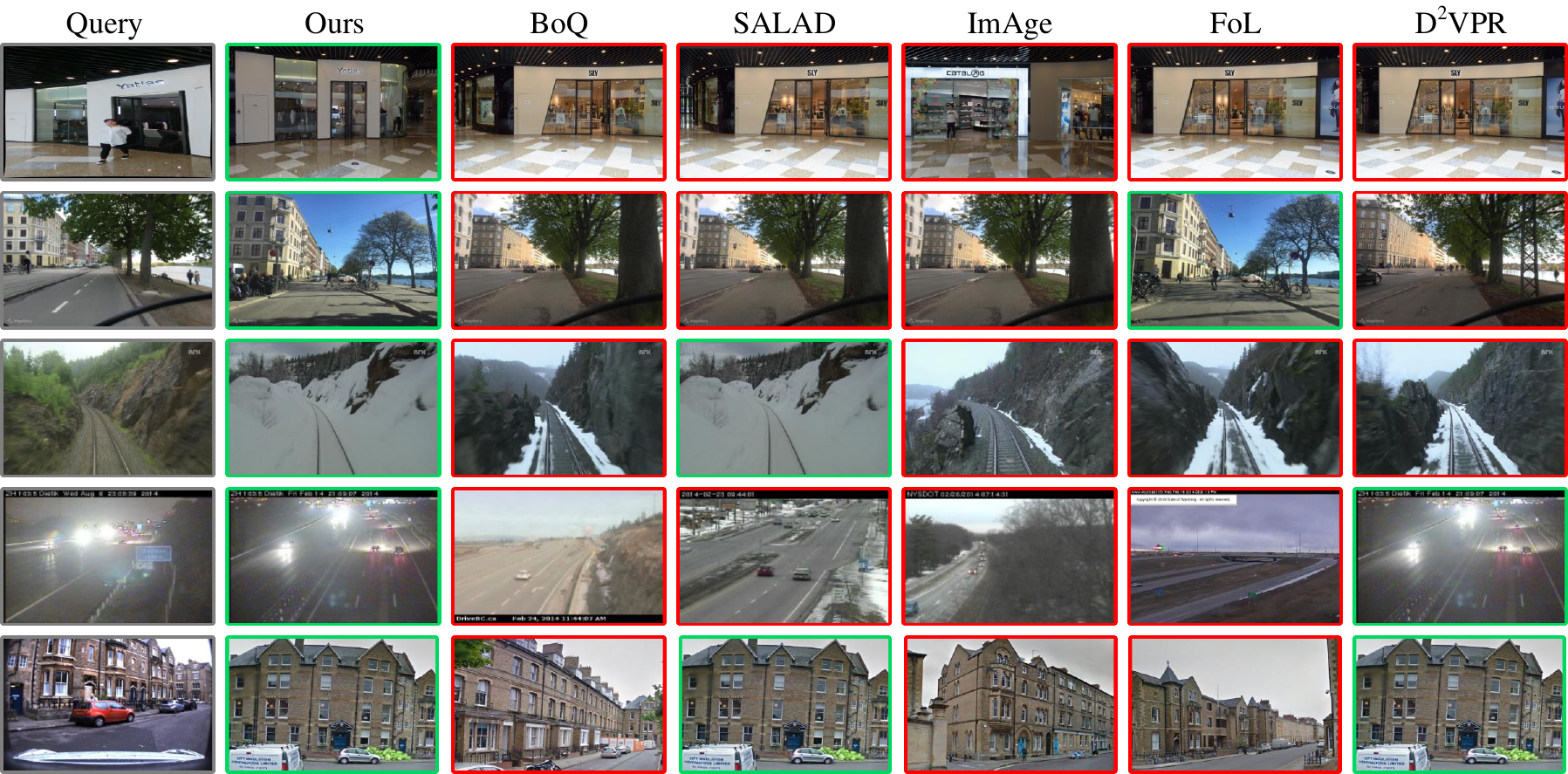}
    \caption{Qualitative VPR results. Each row shows a query and the Top-1
    retrievals of BiSA and competing methods. Green and red borders denote
    correct and incorrect matches, respectively.}
    \label{fig:supp_vpr}
\end{figure*}

\begin{figure}
    \centering
    \includegraphics[width=1\linewidth]{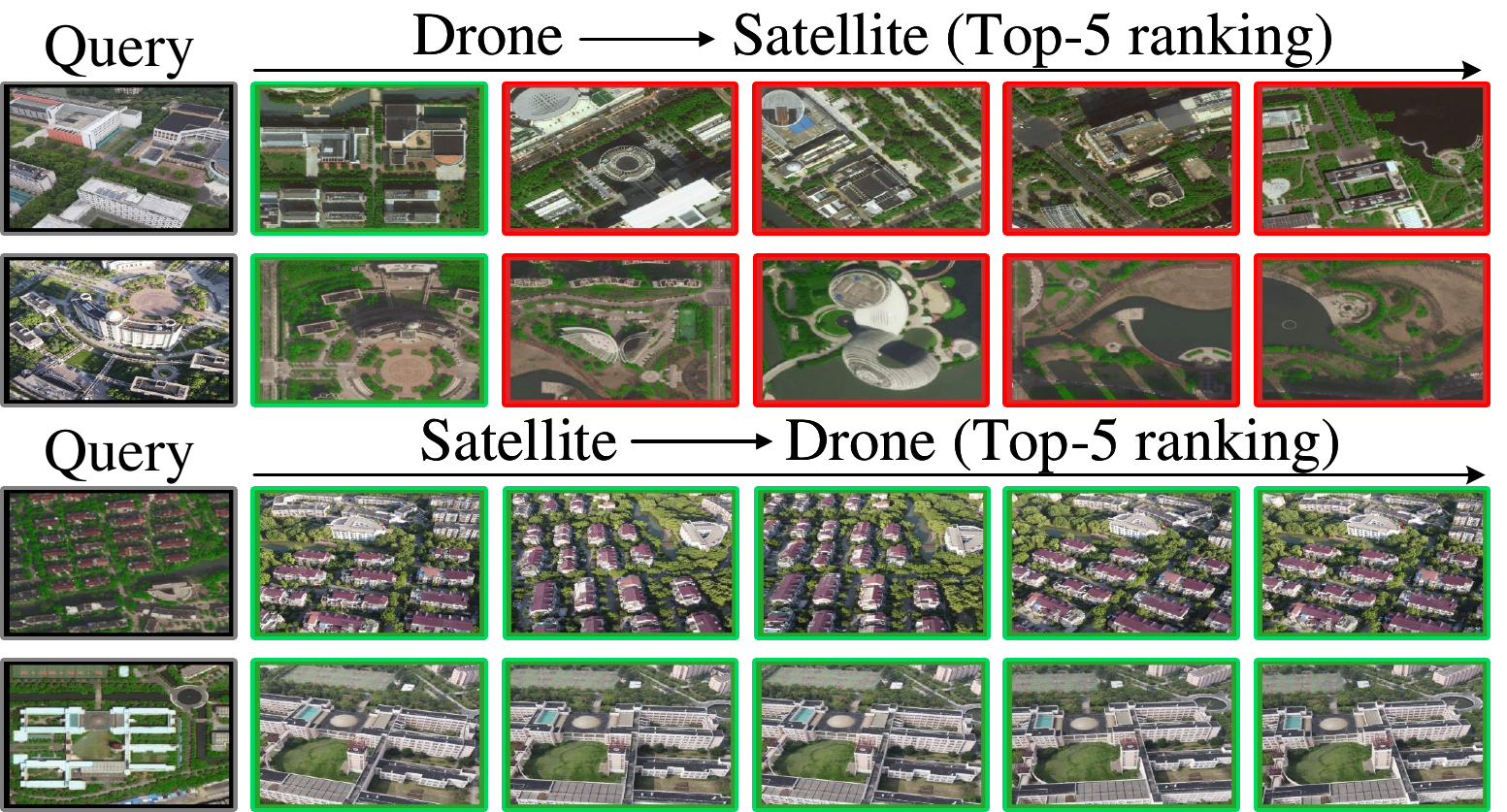}
    \caption{Qualitative CVGL results. Top-5 Drone$\to$Satellite (upper) and
    Top-5 Satellite$\to$Drone (lower) rankings produced by BiSA. Green and
    red borders denote correct and incorrect matches, respectively.}
    \label{fig:supp_cvgl}
\end{figure}

\begin{figure}
    \centering
    \includegraphics[width=1\linewidth]{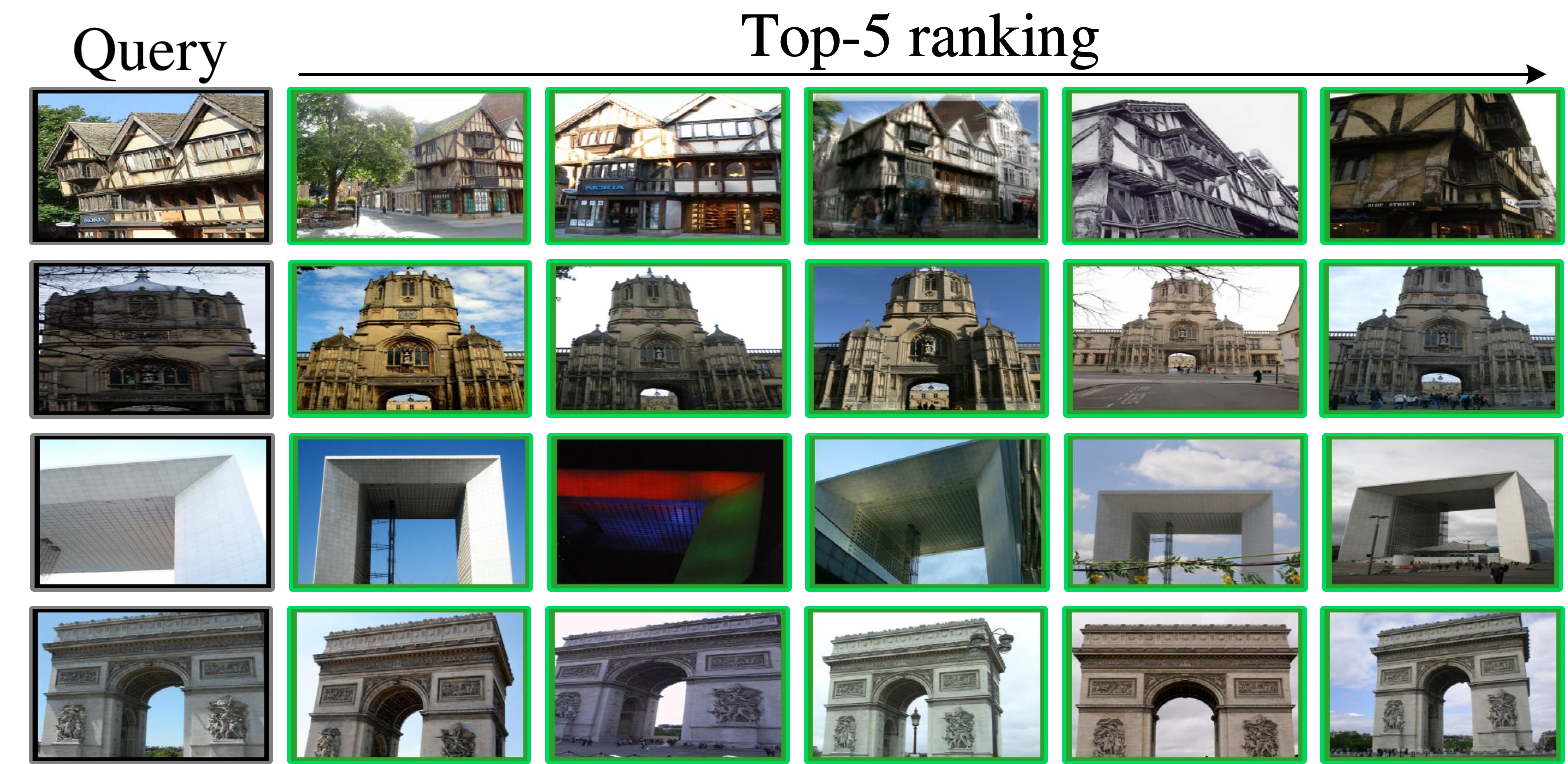}
    \caption{Qualitative ILIR results. Each row shows a query and its Top-5
    retrievals produced by BiSA. Green borders indicate correct matches.}
    \label{fig:supp_ilir}
\end{figure}

Complementing the qualitative analysis in the main paper, we provide
additional retrieval visualizations for VPR
(Figure~\ref{fig:supp_vpr}), CVGL (Figure~\ref{fig:supp_cvgl}), and ILIR
(Figure~\ref{fig:supp_ilir}). Green and red borders denote correct and
incorrect retrievals, respectively. For VPR, we compare the Top-1 results of
BiSA with representative competing methods, while the CVGL and ILIR figures
show the Top-$K$ rankings produced by BiSA.

\noindent\textbf{Visual Place Recognition.}

Figure~\ref{fig:supp_vpr} covers five representative VPR conditions:
repetitive indoor storefronts, urban viewpoint variation, severe seasonal
change, day--night appearance variation, and cross-platform domain shift.
BiSA retrieves the correct place in all displayed cases, whereas competing
methods frequently select perceptually similar locations. These examples
highlight the value of coordinating structural and fine-grained evidence.
The competition axis distributes scene content across complementary slots,
reducing redundant responses to repeated regions, while the selection axis
allows different slots to preserve broad layouts or concentrate on sparse
cues such as signs, storefront boundaries, road contours, and railway
geometry. PTF further suppresses unreliable tokens before subsequent CAC iterations, enabling the final descriptor to retain both stable structure
and locally discriminative evidence under substantial visual change.

\noindent\textbf{Cross-View Geo-Localization.}

Figure~\ref{fig:supp_cvgl} presents the zero-shot SUES-200 evaluation in
both retrieval directions. For Drone$\to$Satellite retrieval, BiSA ranks the
correct satellite image first despite distractors with similar road networks,
vegetation patterns, and building arrangements. For
Satellite$\to$Drone retrieval, all displayed Top-5 results correspond to the
query location across different drone viewpoints. This behavior is consistent
with the structural robustness encouraged by BiSA: the competition axis
organizes complementary scene components on a common comparison scale, while
structure-oriented slots preserve large-scale spatial arrangements across
viewpoint and scale changes. The resulting descriptor therefore remains
stable even when the same location appears under substantially different
imaging geometries.

\noindent\textbf{Instance-Level Image Retrieval.}

Figure~\ref{fig:supp_ilir} shows queries from $\mathcal{R}$Oxf and
$\mathcal{R}$Par together with their Top-5 retrievals. All displayed results
preserve the correct landmark identity under substantial viewpoint, scale,
illumination, and cropping changes. Unlike place-level matching, these cases
require fine-grained discrimination based on localized instance-specific
patterns, including timber arrangements, window layouts, arch contours,
dome geometry, and facade ornaments. BiSA obtains this capability through
the slot-specific selection temperatures, which produce detail-focused slots
that concentrate on sparse discriminative evidence. Combined with
competition-induced complementary decomposition, these localized responses
retain multiple distinctive parts of the same landmark rather than relying
on a single global appearance cue. This behavior is consistent with BiSA's
strong performance under the Hard protocol, where visually similar
distractors place greater demands on detail-level discriminability.

\section{Extended Ablation and Sensitivity Analysis}
\label{app:extended_ablation}

Complementing the component ablations reported in the main paper, this
section provides four additional analyses under the VPR setting on Baidu
Mall, Nordland$^{*}$, and MSLS-val: the sensitivity of BiSA to the numbers of slots
and iterations, its generalization across DINOv2 backbones of different
capacities, the effect of the asymmetric temperature design in CAC, and an
analysis of how token features and CAC-derived reliability cues contribute
to PTF.

\subsection{Sensitivity to the Number of Slots and Iterations}

We study the number of slots $P$ and the number of BiSA iterations $T$ on
Baidu Mall, Nordland$^{*}$, and MSLS-val
(Table~\ref{tab:supp_slot_iteration_sensitivity}). In the slot-number study,
each slot retains a 256-dimensional output projection, so the final
descriptor dimension varies with $P$. In the iteration study, we fix $P=20$
and retain the 5120-dimensional descriptor.

\begin{table}[t]
\centering
\setlength{\tabcolsep}{3.0pt}
\renewcommand{\arraystretch}{1.06}
\fitwidth{{\small
\begin{tabular}{
@{}lccccc@{}}
\toprule
\textbf{Setting} &
\textbf{Value} &
\textbf{Baidu M.} &
\textbf{Nordland$^{*}$} &
\textbf{MSLS} &
\textbf{Avg.} \\
\midrule
\multicolumn{6}{c}{\textbf{Number of Slots}} \\
\midrule
$P$ & 8
& 72.8 & 82.3 & 92.4 & 82.5 \\

$P$ & 12
& 73.7 & 83.5 & 93.1 & 83.4 \\

$\boldsymbol{P}$ & \textbf{20}
& \textbf{74.8}
& \textbf{84.8}
& \textbf{93.8}
& \textbf{84.5} \\

$P$ & 32
& 74.5 & 84.4 & 93.6 & 84.2 \\
\midrule
\multicolumn{6}{c}{\textbf{Number of Iterations}} \\
\midrule
$T$ & 1
& 71.9 & 81.0 & 91.6 & 81.5 \\

$T$ & 2
& 73.9 & 83.7 & 93.0 & 83.5 \\

$\boldsymbol{T}$ & \textbf{3}
& \textbf{74.8}
& \textbf{84.8}
& \textbf{93.8}
& \textbf{84.5} \\

$T$ & 4
& 74.6 & 84.6 & 93.6 & 84.3 \\
\bottomrule
\end{tabular}
}}
\caption{Sensitivity to the numbers of slots and iterations. Results are
reported as R@1 (\%). The default configuration is $P=20$ and $T=3$.}
\label{tab:supp_slot_iteration_sensitivity}
\end{table}

Increasing the number of slots from 8 to 20 steadily improves performance,
indicating that additional slots provide sufficient capacity for
non-redundant scene decomposition and slot specialization across the
structure--detail spectrum. Increasing
$P$ to 32 yields no further improvement, suggesting diminishing returns as
the slot set becomes redundant.

Increasing the number of iterations from one to three also produces consistent
gains. In particular, the weaker result at $T=1$ confirms that a one-step
aggregation process is insufficient to fully coordinate structural and fine-grained evidence. Performance largely saturates after three iterations, while
a fourth iteration provides no meaningful improvement. We therefore use
$P=20$ and $T=3$ as the default configuration.

\subsection{Generalization across Backbones}
\label{app:backbone_generalization}

We evaluate BiSA with DINOv2-S, DINOv2-B, and DINOv2-L to examine its
generalization across different backbone capacities
(Table~\ref{tab:supp_backbone_generalization}). All variants use $P=20$
slots, $T=3$ iterations, a 512-dimensional aggregation space, and a
5120-dimensional global descriptor. Only the final three Transformer blocks
are fine-tuned, while the remaining training settings follow the VPR protocol
used in the main paper.

\begin{table}[t]
\centering
\setlength{\tabcolsep}{3.4pt}
\renewcommand{\arraystretch}{1.06}
\fitwidth{{\small
\begin{tabular}{@{}lcccc@{}}
\toprule
\textbf{Backbone} &
\textbf{Baidu M.} &
\textbf{Nordland$^{*}$} &
\textbf{MSLS} &
\textbf{Avg.} \\
\midrule
DINOv2-S
& 72.3
& 82.1
& 92.2
& 82.2 \\

DINOv2-B
& 74.8
& 84.8
& 93.8
& 84.5 \\

DINOv2-L
& \textbf{75.3}
& \textbf{85.3}
& \textbf{94.2}
& \textbf{84.9} \\
\bottomrule
\end{tabular}
}}
\caption{Generalization of BiSA across DINOv2 backbones. Results are reported
as R@1 (\%).}
\label{tab:supp_backbone_generalization}
\end{table}

BiSA maintains strong retrieval performance across all three backbone
capacities, indicating that its aggregation mechanism is not restricted to a
specific DINOv2 model size. DINOv2-S remains competitive despite its lower
capacity, while scaling from DINOv2-B to DINOv2-L provides only modest gains.
This suggests that BiSA's gains arise primarily from coordinating evidence
across the structure--detail spectrum, rather than from increased backbone
capacity alone.

\subsection{Effect of Asymmetric Temperature Design}
\label{app:temperature_ablation}

We study the asymmetric temperature design of CAC by treating its two axes as
independent factors and further compare learned selection temperatures with
fixed alternatives (Table~\ref{tab:supp_temperature_ablation}). In the
default parameterization, the competition axis uses a single shared
temperature $\tau_c$, while the selection axis assigns each slot its own
$\tau_{s,p}$. All variants keep $P=20$, $T=3$, PTF enabled, the clipping
range $[\tau_{\min},\tau_{\max}]=[0.05,1.0]$, and the VPR training protocol
of the main paper, such that configuration~(e) reproduces Full BiSA.
$\Delta H=\max_p\bar{H}_p-\min_p\bar{H}_p$ denotes the per-slot entropy
spread in nats, where $\bar{H}_p$ is defined in Eq.~5 of the main paper. It
serves as a proxy for how broadly the slots span the structure--detail
spectrum.

\begin{table*}[t]
\centering
\setlength{\tabcolsep}{5.0pt}
\renewcommand{\arraystretch}{1.08}
\fitwidth{{\small
\begin{tabular}{@{}lcccccccc@{}}
\toprule
\textbf{Configuration} &
$\boldsymbol{\tau_c}$ &
$\boldsymbol{\tau_s}$ &
\textbf{\#Temp.} &
\textbf{Baidu Mall} &
\textbf{Nordland$^{*}$} &
\textbf{MSLS-val} &
\textbf{Avg.} &
$\boldsymbol{\Delta H}$ \\
\midrule
\multicolumn{9}{c}{\textbf{Placement of the slot-specific temperatures}} \\
\midrule
(a) Tied across both axes
& \multicolumn{2}{c}{tied} & 1
& 72.9 & 82.1 & 92.2 & 82.4 & 0.21 \\

(b) Shared within each axis
& shared & shared & 2
& 73.2 & 82.6 & 92.6 & 82.8 & 0.24 \\

(c) Reversed asymmetry
& slot & shared & $P{+}1$
& 72.4 & 81.7 & 91.9 & 82.0 & 0.31 \\

(d) Slot-specific on both axes
& slot & slot & $2P$
& 73.7 & 83.2 & 93.0 & 83.3 & 1.48 \\

\textbf{(e) Ours}
& shared & slot & $P{+}1$
& \textbf{74.8} & \textbf{84.8} & \textbf{93.8}
& \textbf{84.5} & \textbf{1.62} \\
\midrule
\multicolumn{9}{c}{\textbf{Learned vs.\ fixed selection temperatures}} \\
\midrule
(f) Fixed $\tau_{s,p}=0.1$
& shared & fixed & 1
& 72.1 & 81.4 & 91.6 & 81.7 & 0.18 \\

(g) Fixed $\tau_{s,p}=0.5$
& shared & fixed & 1
& 72.8 & 82.0 & 92.2 & 82.3 & 0.22 \\

(h) Fixed $\tau_{s,p}=1.0$
& shared & fixed & 1
& 72.3 & 81.8 & 92.0 & 82.0 & 0.19 \\

(i) Fixed, linearly spaced in $[0.1,1.0]$
& shared & fixed & 1
& 73.8 & 83.3 & 93.1 & 83.4 & 1.35 \\
\bottomrule
\end{tabular}
}}
\caption{Ablation of the asymmetric temperature design, with all other
components of Full BiSA fixed. ``shared'' denotes one temperature shared
across slots, ``slot'' denotes a slot-specific temperature, and ``fixed''
denotes a non-learnable value. \#Temp.\ is the number of learnable
temperature parameters; configurations~(c) and~(e) have identical parameter
counts and differ only in the direction of the asymmetry. $\Delta H$ is the
per-slot entropy spread derived from the response entropy defined in Eq.~5
of the main paper and is measured in nats. Results are R@1 (\%).}
\label{tab:supp_temperature_ablation}
\end{table*}

Configurations~(a) and~(b) show that learning separate global temperatures
for the two axes provides only a limited benefit. Configuration~(b) improves
average R@1 by $0.4$ points over the fully tied baseline, while the entropy
spread remains narrow because all slots still share the same selection
temperature.

Configurations~(c) and~(e) use the same number of learnable temperature
parameters but place the slot-specific temperatures on opposite axes. Their
$2.5$-point gap therefore isolates the direction of the asymmetry rather than
added capacity. Slot-specific competition temperatures weaken the common
scale needed for coherent token ownership, whereas the proposed design
retains a shared competition temperature and uses slot-specific selection
temperatures to diversify slot responses across the structure--detail
spectrum. Applying slot-specific temperatures to both axes in~(d) produces a
broad entropy spread but remains $1.2$ points below~(e), further supporting
the proposed asymmetric placement.

Configurations~(f)--(h) show that a single fixed selection temperature cannot
adapt the response granularity of individual slots, resulting in narrow
entropy spreads and average R@1 values $2.2$--$2.8$ points below~(e).
Linearly spaced fixed temperatures in~(i) recover part of the performance by
covering multiple granularities, but remain $1.1$ points behind the learned
design. These results show that both spanning the structure--detail spectrum
and learning its slot-wise allocation are important.
\subsection{Ablation of PTF Reliability Inputs}
\label{app:ptf_input_ablation}

We further examine whether the gains of PTF arise from the CAC-derived
reliability cues or merely from a generic feature-based gating network.
Specifically, we ablate the position-enhanced token feature $X_e$, the
competition claim $c^{(t)}$, and the selection salience $s^{(t)}$
(Table~\ref{tab:supp_ptf_input_ablation}). All filtering variants use the
same progressive mask accumulation as Full BiSA; only the inputs to the
retention predictor are changed. The no-filtering and full-input rows
reproduce the corresponding results in the main paper.

\begin{table}[t]
\centering
\setlength{\tabcolsep}{3.4pt}
\renewcommand{\arraystretch}{1.06}
\fitwidth{{\small
\begin{tabular}{@{}lcccc@{}}
\toprule
\textbf{PTF Input} &
\textbf{Baidu M.} &
\textbf{Nordland$^{*}$} &
\textbf{MSLS} &
\textbf{Avg.} \\
\midrule
No filtering
& 73.1 & 83.1 & 92.2 & 82.8 \\

$X_e$ only
& 73.4 & 83.4 & 92.6 & 83.1 \\

$X_e+c^{(t)}$
& 74.1 & 84.3 & 93.2 & 83.9 \\

$X_e+s^{(t)}$
& 74.3 & 83.8 & 93.0 & 83.7 \\

\textbf{$X_e+c^{(t)}+s^{(t)}$ (Full PTF)}
& \textbf{74.8}
& \textbf{84.8}
& \textbf{93.8}
& \textbf{84.5} \\
\bottomrule
\end{tabular}
}}
\caption{Ablation of the inputs to the PTF retention predictor. All filtering
variants use progressive mask accumulation, and results are reported as
R@1 (\%). Here, $c^{(t)}$ and $s^{(t)}$ denote claim strength under competition and salience under selection at iteration $t$, respectively.}
\label{tab:supp_ptf_input_ablation}
\end{table}

Using $X_e$ alone slightly improves average R@1 from 82.8 to 83.1,
showing that token features provide a basic signal for reliability
estimation. Adding either $c^{(t)}$ or $s^{(t)}$ yields larger gains,
indicating that assignment decisiveness under competition and within-slot
salience provide reliability information beyond token features alone.
Combining both cues achieves the best performance, improving average R@1
by 1.4 points over feature-only gating and by 1.7 points over no filtering.
These results confirm that the two CAC-derived reliability cues are
complementary.

\end{document}